\documentclass{article}

\usepackage{times}
\usepackage{natbib}

\usepackage{url}
\usepackage[section]{placeins}
\usepackage{physics}
\usepackage{graphicx,import}
\usepackage{amsmath}

\usepackage{algorithm}
\usepackage{algorithmic}

\usepackage{hyperref}

\usepackage[accepted]{icml2017}

\icmltitlerunning{The Incredible Shrinking Neural Network}

\newcommand{\Out}[2]{o_{#1}^{(#2)}}
\newcommand{\Target}[1]{t_{#1}}
\newcommand{\Input}[2]{x_{#1}^{(#2)}}
\newcommand{\Weight}[3]{w_{#1#2}^{(#3)}}
\newcommand{\Con}[3]{c_{#1#2}^{(#3)}}

\begin{document} 

\twocolumn[
\icmltitle{The Incredible Shrinking Neural Network: New Perspectives on Learning Representations through the Lens of Pruning}

\icmlauthor{Aditya Sharma}{adityas2@alumni.cmu.edu}
\icmlauthor{Nikolas Wolfe}{nwolfe@cs.cmu.edu}
\icmlauthor{Bhiksha Raj}{bhiksha@cs.cmu.edu}
\icmladdress{Carnegie Mellon University, 5000 Forbes Avenue, Pittsburgh, PA 15213 USA}

\icmlkeywords{pruning neural networks, learning representations, deep learning}

\vskip 0.3in
]

\begin{abstract}
How much can pruning algorithms teach us about the fundamentals of learning representations in neural networks? And how much can these fundamentals help while devising new pruning techniques? A lot, it turns out. Neural network pruning has become a topic of great interest in recent years, and many different techniques have been proposed to address this problem. The decision of what to prune and when to prune necessarily forces us to confront our assumptions about how neural networks actually learn to represent patterns in data. In this work, we set out to test several long-held hypotheses about neural network learning representations, approaches to pruning and the relevance of one in the context of the other. To accomplish this, we argue in favor of pruning whole neurons as opposed to the traditional method of pruning weights from optimally trained networks. We first review the historical literature, point out some common assumptions it makes, and propose methods to demonstrate the inherent flaws in these assumptions. We then propose our novel approach to pruning and set about analyzing the quality of the decisions it makes. Our analysis led us to question the validity of many widely-held assumptions behind pruning algorithms and the trade-offs we often make in the interest of reducing computational complexity. We discovered that there is a straightforward way, however expensive, to serially prune 40-70\% of the neurons in a trained network with minimal effect on the learning representation and without any re-training. It is to be noted here that the motivation behind this work is not to propose an algorithm that would outperform all existing methods, but to shed light on what some inherent flaws in these methods can teach us about learning representations and how this can lead us to superior pruning techniques.
\end{abstract}

\section{Introduction and Literature Review}\label{sec1}

Representation of Learning has always been a topic of curiosity in the world of Artificial Neural Networks. It is something that certainly warrants more attention and analysis than it currently receives. It seems very intuitive that the knowledge of how learning is represented in a network of neurons should prove to be useful in developing algorithms. Unfortunately a vast majority of work done in the highly popular fields of Artificial Neural Networks and Deep Learning doesn't focus on leveraging this knowledge. One category of algorithms that can certainly leverage this information and yet surprisingly hasn't is pruning techniques. Neural Network Pruning gained some popularity in the early 90s and after a period of dormancy has recently regained some momentum. In the current work we take a deep dive into existing literature on the topic, both historical and modern, and recognize some areas of failure arising from fundamentally flawed assumptions about learning representations. We then argue in favor of a novel approach to pruning and demonstrate through experiments the superiority of this new way of thinking. We would like to note here that this work does not aim to propose an exact new superior pruning algorithm, but a new way of approaching the subject which isn't flawed in its assumptions and hence promises to perform better, which we empirically demonstrate. 

We begin our exploratory journey in 1989, jumping off from an insightful though largely forgotten observation by \cite{mozer1989skeletonization} concerning the nature of neural network learning. This observation is best summarized in a quotation from \cite{segee1991fault} on the notion of fault-tolerance in multi-layer perceptron networks:

\begin{quotation}
Contrary to the belief widely held, multi-layer networks are \textit{not} inherently fault tolerant. In fact, the loss of a single weight is frequently sufficient to completely disrupt a learned function approximation. Furthermore, having a large number of weights \textit{does not seem} to improve fault tolerance. [Emphasis added]
\end{quotation}

Essentially, \cite{mozer1989using} observed that during training, neural networks do \textit{not} distribute the learning representation evenly or equitably across hidden units. By making this observation alone, this work put itself miles ahead of a lot of work in pruning that followed in the coming two decades. It was a very straightforward observation and yet raised a lot of interesting questions, many of which remain unanswered even today. It largely talked about the \textit{relevance} of individual neurons in a network and argued that it is only a few, elite neurons that learn an approximation of the input-output function, and hence the remaining useless neurons could be trimmed to produce a \textit{skeleton network}. They however, do not go into the details of \textit{why} this happens or \textit{what} exactly is the role of these other noisy neurons, if any. We will go a little deeper into this discussion later in this work.

\subsection{Compressing Neural Networks}
Pruning algorithms, as comprehensively surveyed by \cite{reed1993pruning}, are a useful set of heuristics designed to identify and remove elements from a neural network which are either redundant or do not significantly contribute to the output of the network. This is motivated by the observed tendency of neural networks to overfit to the idiosyncrasies of their training data given too many trainable parameters or too few input patterns from which to generalize, as stated by \cite{chauvin1990generalization}. \cite{baum1989size} demonstrate that there is no way to precisely determine the appropriate size of a neural network a priori given any random set of training instances. Using too few neurons seems to inhibit learning, and so while using pruning techniques in practice it is common to attempt to over-parameterize networks initially using a large number of hidden units and weights, and then prune or compress these fully-trained networks as necessary. Of course, as the old saying goes, there's more than one way to skin a neural network. 

The idea of Skeletonization inspired others to look into the world of faster and better generalization that pruning techniques promised. However in doing so, they somehow lost the essence of the most important result put forward by \cite{mozer1989skeletonization} regarding learning representations and instead focused on improving generalization and memory usage through fast and easy pruning of network elements. Optimal Brain Damage \cite{lecun1989optimal} was one such algorithm that quickly became very popular, as did its later variants such as Optimal Brain Surgeon \cite{hassibi1993second}. These algorithms took a weight centric pruning approach which quickly became the de facto standard for almost all the work that followed, while the original ideas put forward by \cite{mozer1989skeletonization} were largely forgotten. 

In the years to come, pruning algorithms paved the way for model compression algorithms. Model compression often has benefits with respect to generalization performance and the portability of neural networks to operate in memory-constrained or embedded environments. Without explicitly removing parameters from the network, weight quantization allows for a reduction in the number of bytes used to represent each weight parameter, as investigated by \cite{balzer1991weight}, \cite{dundar1994effects}, and \cite{hoehfeld1992learning}. 

A recently proposed method for compressing recurrent neural networks \cite{prabhavalkar2016compression} uses the singular values of a trained weight matrix as basis vectors from which to derive a compressed hidden layer. \cite{Anders2016quant} successfully implemented network compression through weight quantization with an encoding step while others such as \cite{deepcompression2016} have tried to expand on this by adding weight-pruning as a preceding step to quantization and encoding. 

Since \cite{mozer1989skeletonization}, other heuristics affecting the generalization behavior of neural networks have been well studied too, and have been used to avoid overfitting, such as dropout \cite{srivastava2014dropout}, maxout \cite{goodfellow2013maxout}, and cascade correlation \cite{fahlman1989cascade}, among others. Of course, while cascade correlation specifically tries to construct minimal networks, many techniques that try to improve network generalization do not explicitly attempt to reduce the total number of parameters or the memory footprint of a trained network per se.  

In summary, we can see how since the ideas put forth by \cite{mozer1989skeletonization} back in 1989 the research in this area went on a tangential path which led to most of our popular pruning techniques today. In subsequent sections of this paper, we will look deeper into the original ideas from 1989, try to answer some questions that were left unanswered, demonstrate how not leveraging these original ideas introduced fundamental flaws in today's pruning techniques, and finally make a case for using learning representations to drive decision making in pruning algorithms. Through our experiments we not only concretely validate the theory put forth by \cite{mozer1989using} but we also successfully build on it to show that it is possible to prune networks to 40 to 70 \% of their original size without any major loss in performance.

\section{Pruning Neurons to Shrink Networks}\label{sec2}

\subsection{Why Prune Neurons Instead of Pruning Weights?}

It is surprising that despite the promising ideas presented by \cite{mozer1989skeletonization}
regarding the role of individual neurons in the overall network performance, it was weight pruning that became popular instead. The more one thinks about it, the more one begins to realize why pruning neurons sounds more logical than pruning weights. We look into some of these reasons here.

First is the original argument itself made by \cite{mozer1989skeletonization}. Not all neurons are equal, some of them contribute more than the others, so it makes logical sense to free up space by completely getting rid of these unimportant neurons. Contrast this with eliminating weights, where the pruning criteria relies on some function of actual weight values. At no point during weight-pruning you are getting rid of whole neurons. One might argue that unimportant neurons will end up being removed automatically in a weight-pruning approach as their weights will have low values, but it was demonstrated by \cite{mozer1989using} that this is very much not the case. The actual values of the weights emanating from a neuron have nothing to do with its contribution to the overall performance of the network. Even if a weight-pruning algorithm managed to completely remove a neuron, it might end up being an important neuron which would disrupt the network completely. 

Second, again a quite obvious reason, pruning neurons as opposed to individual weights frees up more space since you eliminate all incoming and outgoing weights from the pruned neuron in one go. The removal of a single weight from a large network is a drop in the bucket in terms of reducing a network's core memory footprint. Somewhat tied to this is the third reason, pruning neurons instead of weights means that you have far fewer (orders of magnitude less) number of elements to consider during the pruning process itself, which makes it much cheaper computationally. Pruning neurons has a practical advantage in terms of quickly reaching a hypothetical target reduction in memory consumption.

Fourth, as touched upon in the first reason, the values of individual weights have nothing to do with the relevance of the corresponding neurons. The neurons represent learning, the weights are just a means of \textit{storing} this learning in memory by splitting it into smaller pieces used to interact with other neurons. Weights provide little information about learning individually and it is their interaction with connected neurons that makes their existence meaningful, a fact that most weight-pruning algorithms tend to overlook. In fact, this is a major drawback in the case of quantization-based algorithms as these methods completely disregard the individual role of neurons and focus solely on quantizing weight values.

Lastly, if you look at it in terms of features, each neuron learns exactly one feature, with the features becoming more complex as you move to layers closer to the output. As in any Machine Learning problem, many of these features are redundant, the ``Train Problem" discussed in \cite{mozer1989using} is a very good basic example of this. In practice, these redundant features should be removed (which is usually done in a ``feature engineering" step in classic ML techniques) as they add noise, increase training time and lead to wrong generalization. In neural networks, which have a more end-to-end architecture as compared to classic ML techniques, this can only be achieved by removing whole neurons. In contrast, pruning weights only \textit{modifies} such features thereby removing their redundant components, but doesn't take into account the overall relevance of the feature and hence never completely removes it.

\subsection{Setting up the experiments}

Now that we have an understanding of how pruning algorithms evolved over the years, some faulty assumptions they made about learning representations, and finally, the benefits of pruning whole neurons instead of individual weights, we are ready to look deeper into the mechanics of popular pruning algorithms. We will look at some other assumptions they make and experimentally analyze their performance when compared to a very simple brute-force based approach which carefully takes into consideration the role of learning representations while pruning networks. Since we have already discussed the benefits of pruning neurons over weights, all experiments presented in this paper will use neuron pruning. We will also not experimentally evaluate algorithms that quantize weights and compress networks because one, they are not exactly \textit{pruning} algorithms and two, their biggest fundamental flaw has already been discussed while making the argument against weight pruning earlier in this section.

It is beyond the scope of this paper to do an empirical analysis of every single pruning algorithm out there. We could have selected a top few popular algorithms for our comparisons but that would defeat the central purpose of this work, which is to look for major flaws in the fundamental assumptions made by these algorithms. This requires an approach that is representative of the central ideas used by most, if not all of these existing algorithms.

\subsubsection{Linear and Quadratic Approximation Approach}
The overall structure of most pruning algorithms is pretty much the same: you rank individual elements (weights or neurons) based on some criteria (usually related to the element's impact on the network error) and keep trimming away the least relevant elements until you see a drop in performance. The overall impact of an element on the network performance in these algorithms is typically represented by a linear or quadratic approximation of change in error. We will compare both these approaches to our simple brute-force approach. The experiments presented in this work will follow a ranking-based procedure very similar to most popular pruning techniques. The linear and quadratic error estimates were derived using a Taylor Series representation of the network error, very similar to what \cite{mozer1989skeletonization} did in their original work. The math involved for the quadratic estimate is somewhat complicated, as is its implementation which includes gradients collected from second-order back-propagation and hence, we will not go into the details of either in the main body of the paper. All these details however, are provided in the Supplementary Material as an Appendix for readers who might be interested in pursuing and implementing this.

\subsubsection{Brute Force Removal Approach}
This is perhaps the most intuitive, naive and yet the most accurate method for pruning a network, but surprisingly, not very popular, perhaps due to its time complexity. It does not make any assumptions about the learning representations in the network, nor does it make any approximate estimates about the impact of individual elements on the network's output, which make it a perfect candidate to serve as the ground truth for our experiments. This method explicitly evaluates each neuron in the network. The idea is to manually check the effect of every single neuron on the output. This is done by running a forward propagation on the validation set $K$ times (where $K$ is the total number of neurons in the network), turning off exactly one neuron each time (keeping all other neurons active) and noting down the change in error.

\subsubsection{Neuron Inter-dependencies}
Something that was briefly touched upon in the Section \ref{sec1} was the fact that while the findings of \cite{mozer1989skeletonization} about learning representations were of great significance, one important question remained unanswered: Do the noisy/unimportant neurons have a role to play once the network is trained? Although there is no empirical evidence to support their role \textit{during} the training process itself, one can still easily imagine the necessity to include these extra neurons to aid better generalization. However, their role \textit{after} the network is trained is something we would like to explore. \cite{mozer1989skeletonization} proved that these neurons don't really contribute to the network output (which is why we prune them), but does their existence impact the other neurons in some way? In order to find out, we begin our experiments exactly where \cite{mozer1989skeletonization} left off, and we don't assume these unnecessary neurons to have any impact on each other or the relevant neurons. We use Algorithm \ref{algo1} to verify how correct this assumption is. We will analyze our observations and follow them up with with further experiments in the next section. The idea here is to generate a single ranked list based on the values of $\Delta E_{k}$, the Taylor Series approximation of the change in the overall error of the network due to the neuron $k$. This involves a single pass of second-order back-propagation (without weight updates) to collect the gradients for each neuron. The neurons from this rank-list (with the lowest values of $\Delta E_{k}$) are then pruned according to the stopping criterion decided.

\begin{algorithm}
 \caption{Single Overall Ranking}
 \label{algo1}
 \begin{algorithmic}
 \STATE{\textbf{Data:} An optimally trained network, training dataset}
 \STATE {\textbf{Output:}A pruned network}
 \STATE Initialize and define stopping criterion;
 
 \STATE Perform forward propagation over the training set;
 
  \STATE Perform second-order back-propagation without updating weights and collect linear and quadratic gradients;
  
  \STATE Rank the remaining neurons based on $\Delta E_{k}$;
  
 \WHILE{stopping criterion is not met}
 \STATE Remove the last ranked neuron;
  
  \ENDWHILE
\end{algorithmic}
\end{algorithm}

\section{Experimental Results \& Discussion}\label{sec3}
For most experiments presented in this section, the MNIST database of Handwritten Digits by \cite{lecun-mnisthandwrittendigit-2010} was used. It is worth noting that due to the time taken by the brute force algorithm we used a 5000 image subset of the MNIST database in which we normalized the pixel values between 0 and 1.0, compressed the image sizes to 20x20 images rather than 28x28, and used a 1 or 2-layer fully connected feedforward neural network with a total of 100 neurons and sigmoid activation, so the starting test accuracy reported here appears higher than those reported by LeCun et al. This modification does not affect the interpretation of the presented results because the fundamental learning problem does not change with a larger dataset or input dimension or network architecture. Also, as has been pointed out multiple times before, the aim of this work is not to propose a superior algorithm, but to point out flaws in the assumptions that existing algorithms make by means of the better performing brute force method, hence the choice of dataset does not impact our results. Wherever necessary, some other toy datasets have also been used to empirically verify some novel theoretical insights presented in the previous sections of this work. The sigmoid activation function was chosen since it makes estimating gradients easy (as compared to a ReLU activation function), and more importantly because it has been the popular choice historically, which helps us make sure our experiments are representative of the older pruning techniques as well.

\subsection{Neuron Inter-dependencies}\label{sec3.1}
\begin{figure}[!ht]
\centering
\includegraphics[width=0.49\linewidth]{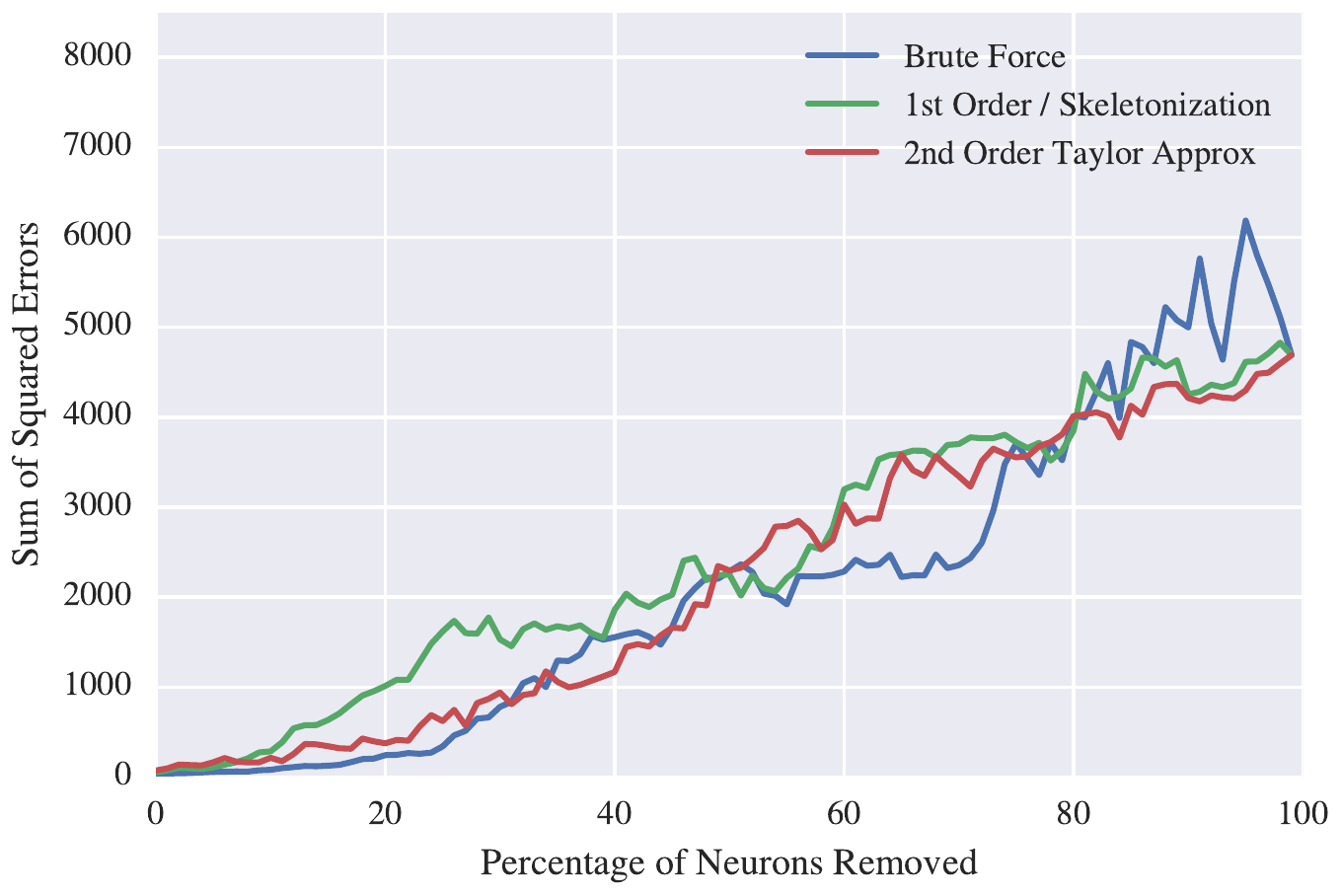}
\includegraphics[width=0.49\linewidth]{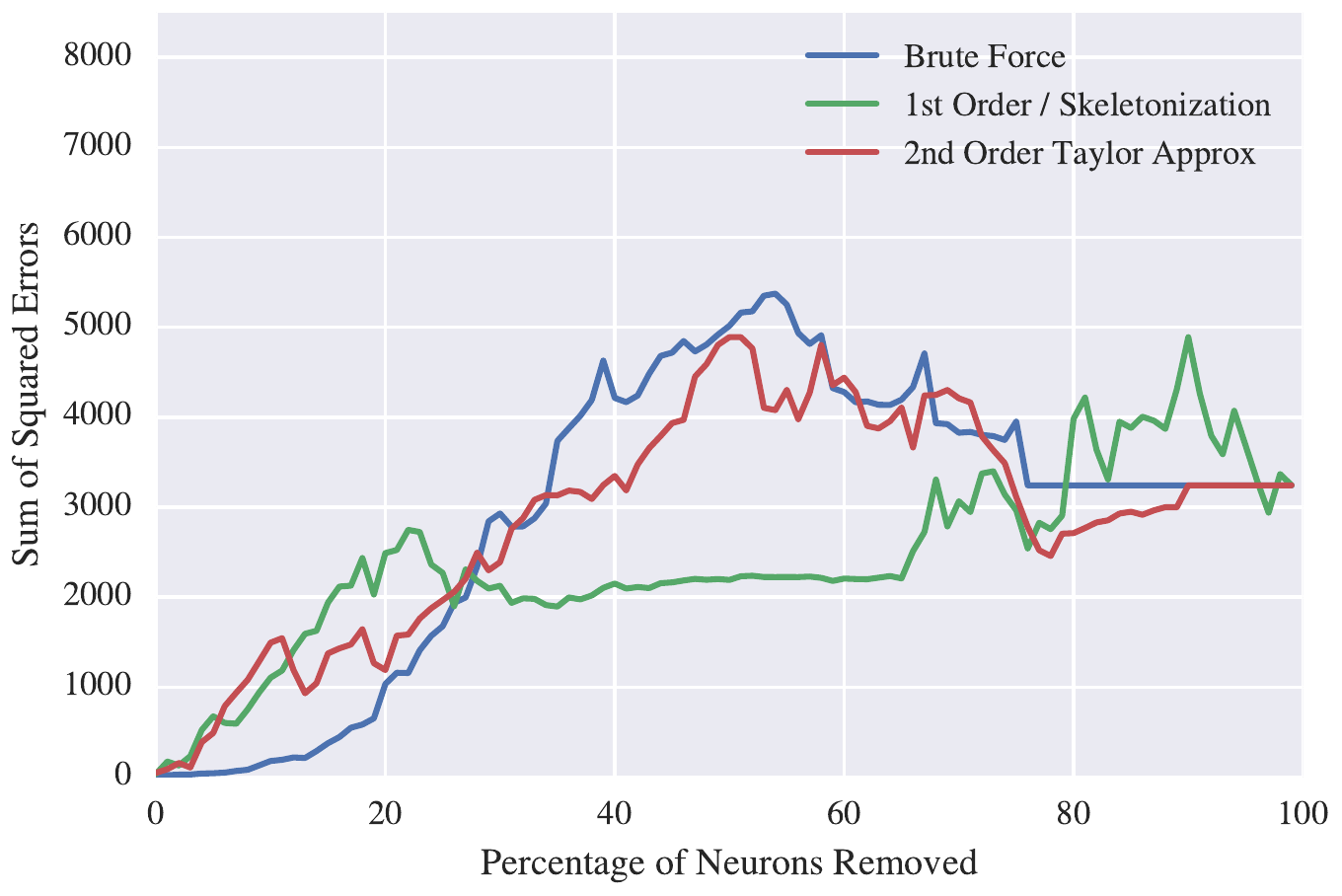}
\caption{Degradation in squared error after pruning a 1-layer network (left) and a 2-layer network (right) using The Single Overall Ranking algorithm (\textbf{Network 1:} 1 layer, 100 neurons, starting test accuracy: 0.998; \textbf{Network 2:} 2 layers, 50 neurons/layer, starting test accuracy: 1.000)}
\label{fig:mnist-single-ranking}
\end{figure}

We start off by trying to answer the question we put forth at the end of the Section \ref{sec2}. Do the irrelevant neurons in a network still have a role to play after the training is completed? As mentioned before, we first start off by assuming that the answer to the question is no, simply because we saw no historical evidence that suggested otherwise. We present the results for a single-layer neural network in Figure \ref{fig:mnist-single-ranking}, using Single Overall Ranking (Algorithm \ref{algo1}) as proposed in Section \ref{sec2}. After training, each neuron is assigned its permanent ranking based on the three criteria discussed previously: A brute force ``ground truth'' ranking, and the linear and quadratic error approximations that represent popular existing techniques in principle.

As can be observed, none of the three approaches perform well at all. The performance starts dropping as soon as neurons are removed according to the ranked list. This is a very important result, although not unexpected. This behavior implies that there is a flaw in our assumption that the irrelevant neurons have no role to play in the network's performance post-training. We performed the same experiment on other toy datasets and we got the same results. These results answer the question posed at the beginning of this section. These \textit{irrelevant} neurons do have a role to play post-training. Their existence is somehow linked to the performance of the relevant neurons, and as \cite{segee1991fault} assert, the removal of any single hidden unit at random can actually result in a network fault. We have empirically, if not numerically, proved the role of these irrelevant units but this makes us wonder if it is possible, without going into the mathematics of the inter-neural relationships, to come up with a way to still prune networks efficiently.

What we did wrong in the experiment above was ignore the fact that after a neuron is removed, no matter how irrelevant its existence was to the overall performance, it \textit{does} have an impact on the performance of the other neurons, some of which might be highly relevant in the overall context. Keeping that, and the intriguing fault tolerance (or lack thereof) of trained neural networks in mind, we decided to modify our algorithm. In this greedy variation of the algorithm (Algorithm \ref{algo2}), after each neuron removal, the remaining network undergoes a single forward and backward pass of second-order back-propagation (without weight updates) and the rank list is formed again. Hence, each removal involves a new pass through the network. This method takes into account the dependencies the neurons might have on one another each time the rank list is re-formed and relies on the low fault tolerance of the network which disallows large error drops that might come with the removal of a neuron, but only as long as all the relevant neurons are still intact.

\begin{algorithm}
\caption{Iterative Re-Ranking}
 \label{algo2}
 \begin{algorithmic}
\STATE{\textbf{Data:} An optimally trained network, training dataset}
 \STATE {\textbf{Output:}A pruned network}
 \STATE Initialize and define stopping criterion;
 
 \WHILE{stopping criterion is not met}
  \STATE Perform forward propagation over the training set;
  
  \STATE Perform second-order back-propagation without updating weights and collect linear and quadratic gradients;
  
  \STATE Rank the remaining neurons based on $\Delta E_{k}$;

  \STATE Remove the worst neuron based on the ranking;
  
 \ENDWHILE
 
\end{algorithmic}
\end{algorithm}

\subsection{Example Regression Problem}\label{sec3.2}
Even though the results above decisively prove the significance of irrelevant neurons, they still do not prove the fact that neuron pruning is actually possible. So far we have considered the results from \cite{mozer1989skeletonization} as the gospel truth. Since these results have been long forgotten, and since most of the reasons cited by us in favor of the novel approach of pruning whole neurons in Section \ref{sec2} were largely theoretical, one might argue for more empirical evidence. Hence, to get some heuristic validation we present results from a simple regression experiment. With this experiment we hope to prove both the existence of non-uniform learning distribution across neurons as well as the efficiency of a neuron-pruning approach.

We trained two networks to learn the cosine function, with one input and one output. This is a task which requires no more than 11 sigmoid neurons to solve entirely, and in this case we don't care about overfitting because the cosine function has a precise definition. Furthermore, the cosine function is a good toy example because it is a smooth continuous function and, as demonstrated by \cite{nielsen2015neural}, if we were to tinker directly with the parameters of the network, we could allocate individual units within the network to be responsible for constrained ranges of inputs, similar to a basis spline function with many control points. This would distribute the learned function approximation evenly across all hidden units, and thus we have presented the network with a problem in which it could productively use as many hidden units as we give it. In this case, a pruning algorithm would observe a fairly consistent increase in error after the removal of each successive unit. In practice however, regardless of the number of experimental trials, this is not what happens. The network will always use 10-11 hidden units and leave the rest to cancel each others influence, as we will see.

\begin{figure}[!ht]
\centering
\includegraphics[width=0.49\linewidth]{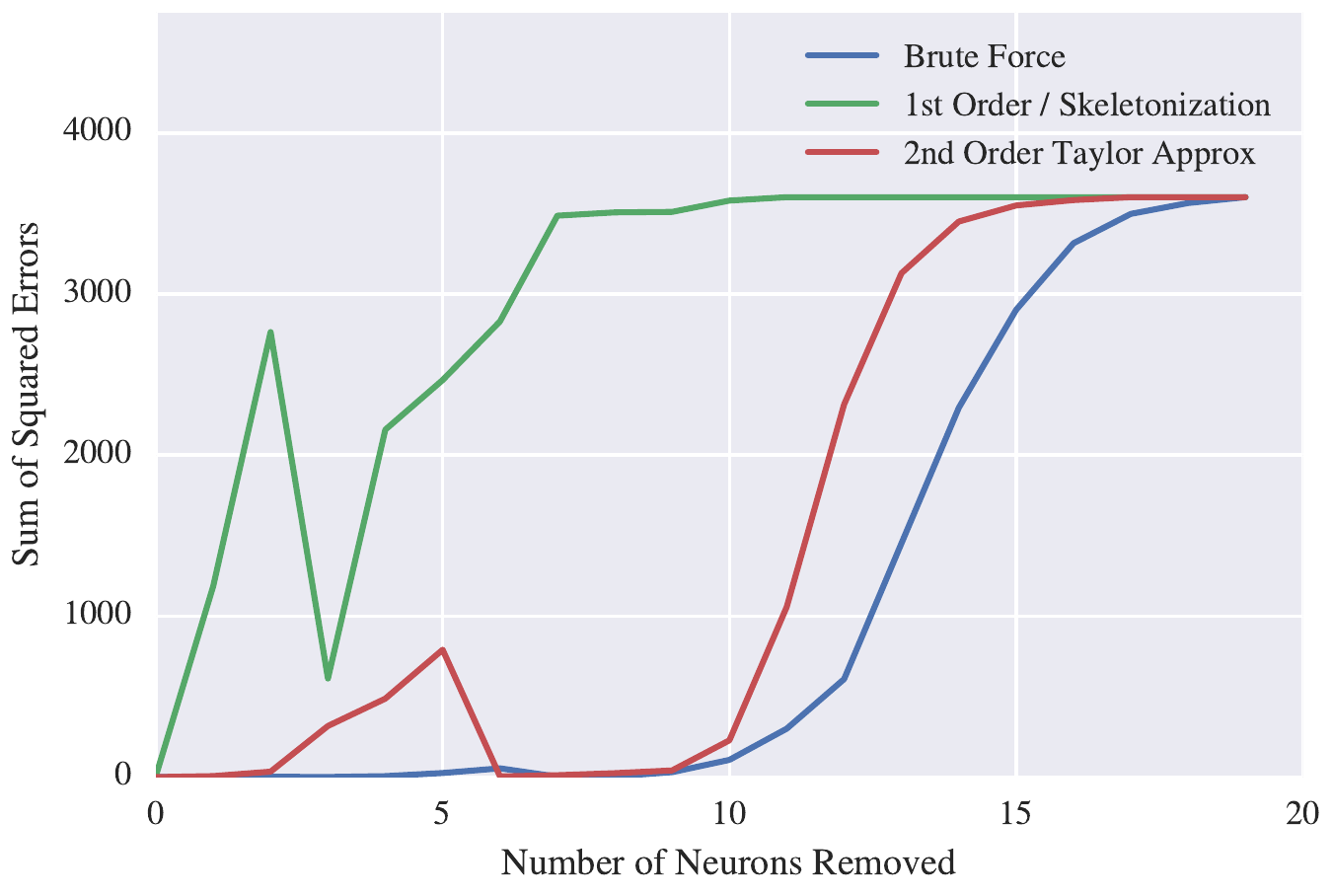}
\includegraphics[width=0.49\linewidth]{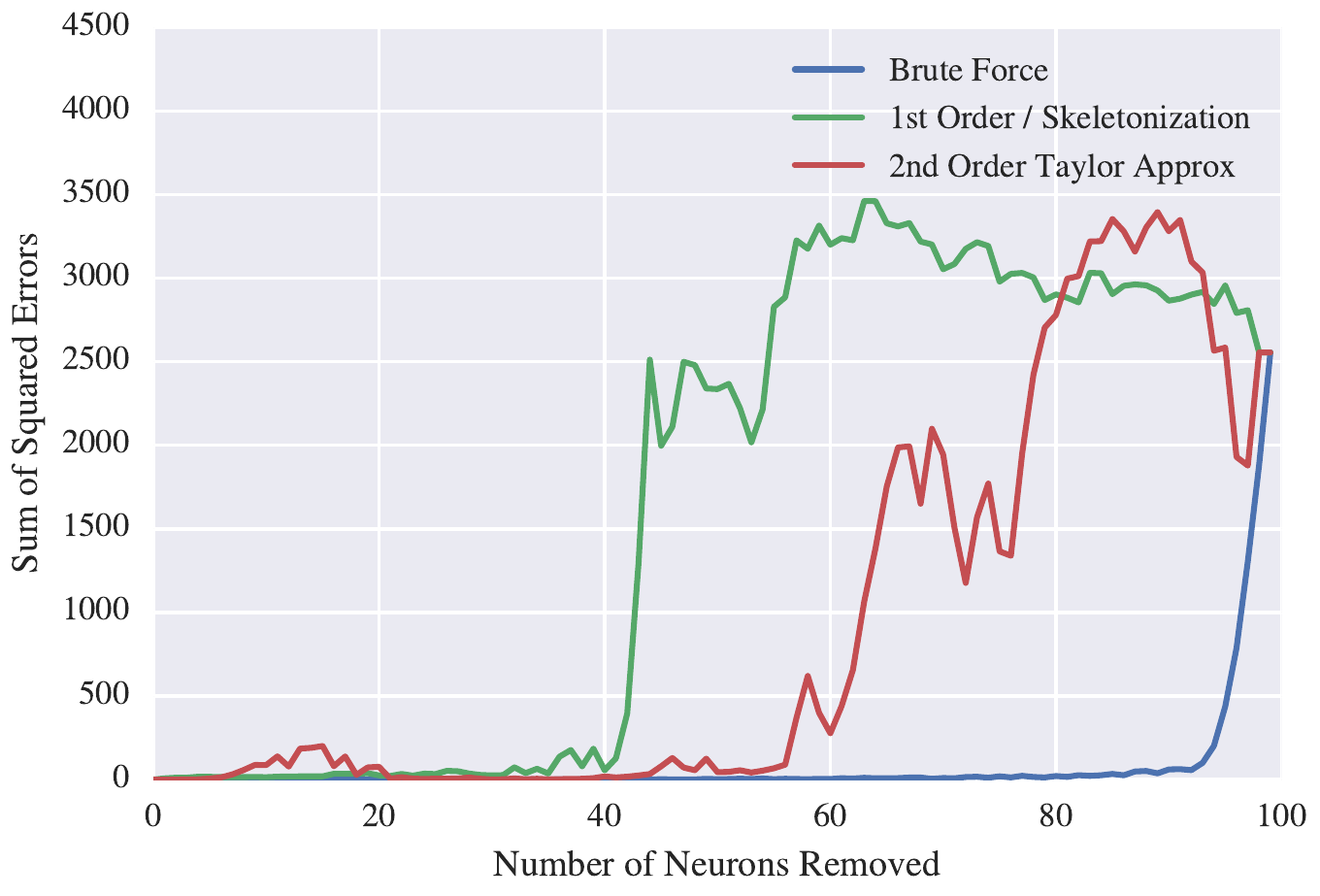}
\caption{Degradation in squared error after pruning a two-layer network trained to compute the cosine function (\textbf{Left Network:} 2 layers, 10 neurons each, starting test accuracy: 0.9999993, \textbf{Right Network:} 2 layers, 50 neurons each, starting test accuracy: 0.9999996)}
\label{fig:cosine-double-layer}
\end{figure}

Figure \ref{fig:cosine-double-layer} shows two graphs. Both graphs demonstrate the use of Algorithm \ref{algo2} and the comparative performance of the brute-force pruning method (in blue), the first order method (in green), and the second order method (in red). The graph on the left shows the performance of these methods on a network with two layers of 10 neurons (20 total), and the graph on the right shows a network with two layers of 50 neurons (100 total). 

In the left graph, we see that the brute-force method shows a graceful degradation, and the error only begins to rise sharply after 50\% of the total neurons have been removed. The error is basically constant up to that point. In the first and second order methods, we see evidence of poor decision making in the sense that both made mistakes early on, which disrupted the output function approximation. The first order method made a large error early on, though we see after a few more neurons were removed this error was corrected somewhat (though it only got worse from there). This is direct evidence of the lack of fault tolerance in a trained neural network once a relevant unit is removed. This phenomenon is even more starkly demonstrated in the second order method. After making a few poor neuron removal decisions in a row, the error signal rose sharply, and then went back to zero after the 6th neuron was removed. This is due to the fact that the neurons it chose to remove were trained to cancel each others' influence within a localized part of the network. After the entire group was eliminated, the approximation returned to normal. This can only happen if the output function approximation is not evenly distributed over the hidden units in a trained network. 

This phenomenon is even more definitely demonstrated in the graph on the right. Here we see the first order method got ``lucky'' in the beginning and made decent decisions up to about the 40th removed neuron. The second order method had a small error in the beginning which it recovered from gracefully and proceeded to pass the 50 neuron point before finally beginning to unravel. The brute force method, in sharp contrast, shows little to no increase in error at all until 90\% of the neurons in the network have been obliterated. Clearly first and second order methods have some value in that they do not make completely arbitrary choices, but the brute force method is far better at this task. 

This experiment gives us some very important results. One, it empirically verifies the fact that learning is not equally or equitably distributed in a neural network. Two, it proves that pruning whole neurons without a loss in performance is actually possible and three, it shows us some early indications of the inefficient nature of popular first and second order error approximation techniques. Clearly, in the case of the brute force or oracle method, which makes no faulty assumptions about the nature of learning representations, up to 90\% of the network can be completely extirpated before the output approximation even begins to show any signs of degradation. We will explore this further in the next sections.

\subsection{Experiments on MNIST and Toy Datasets}\label{sec3.3}

\begin{figure}[!ht]
\centering
\includegraphics[width=0.49\linewidth]{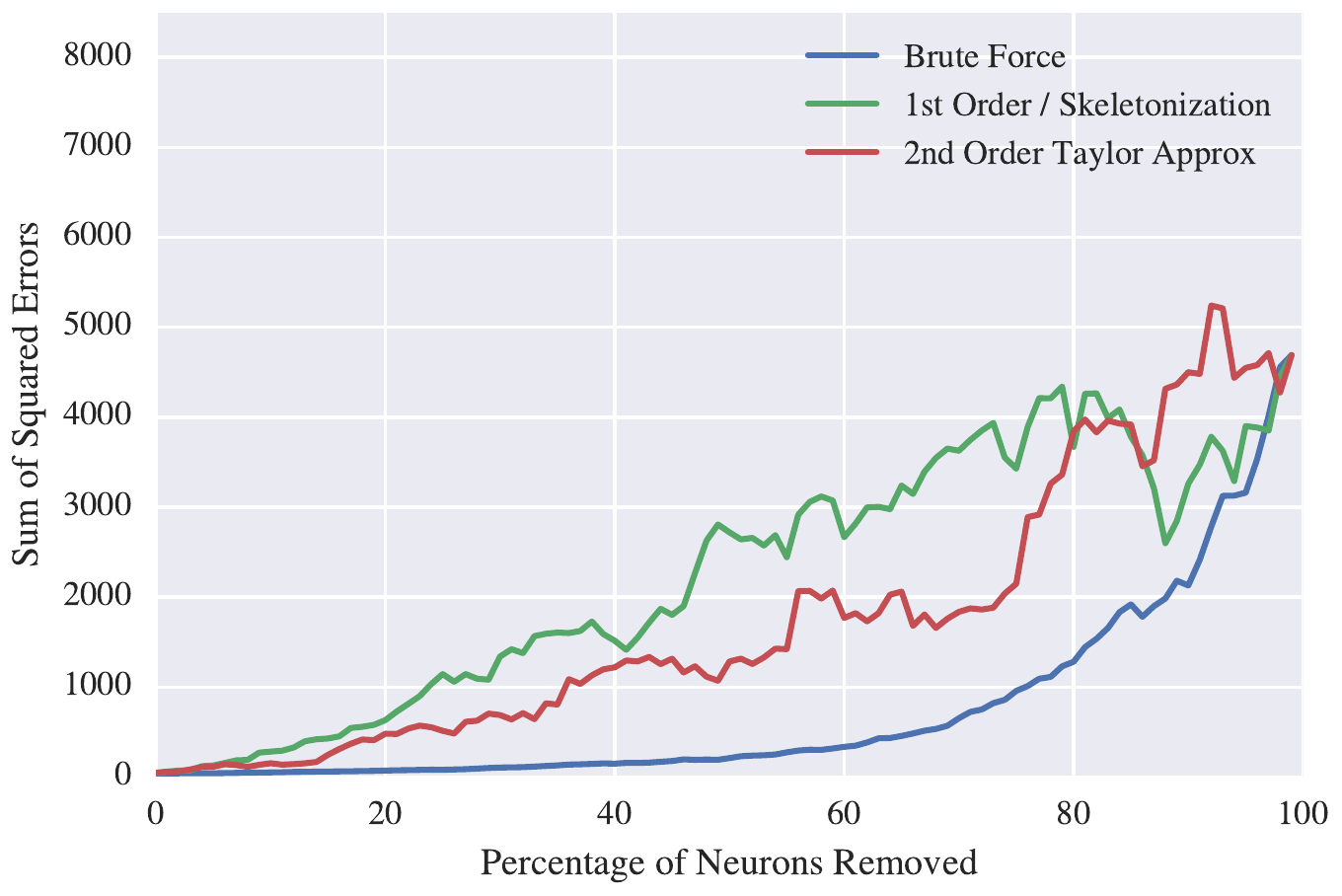}
\includegraphics[width=0.49\linewidth]{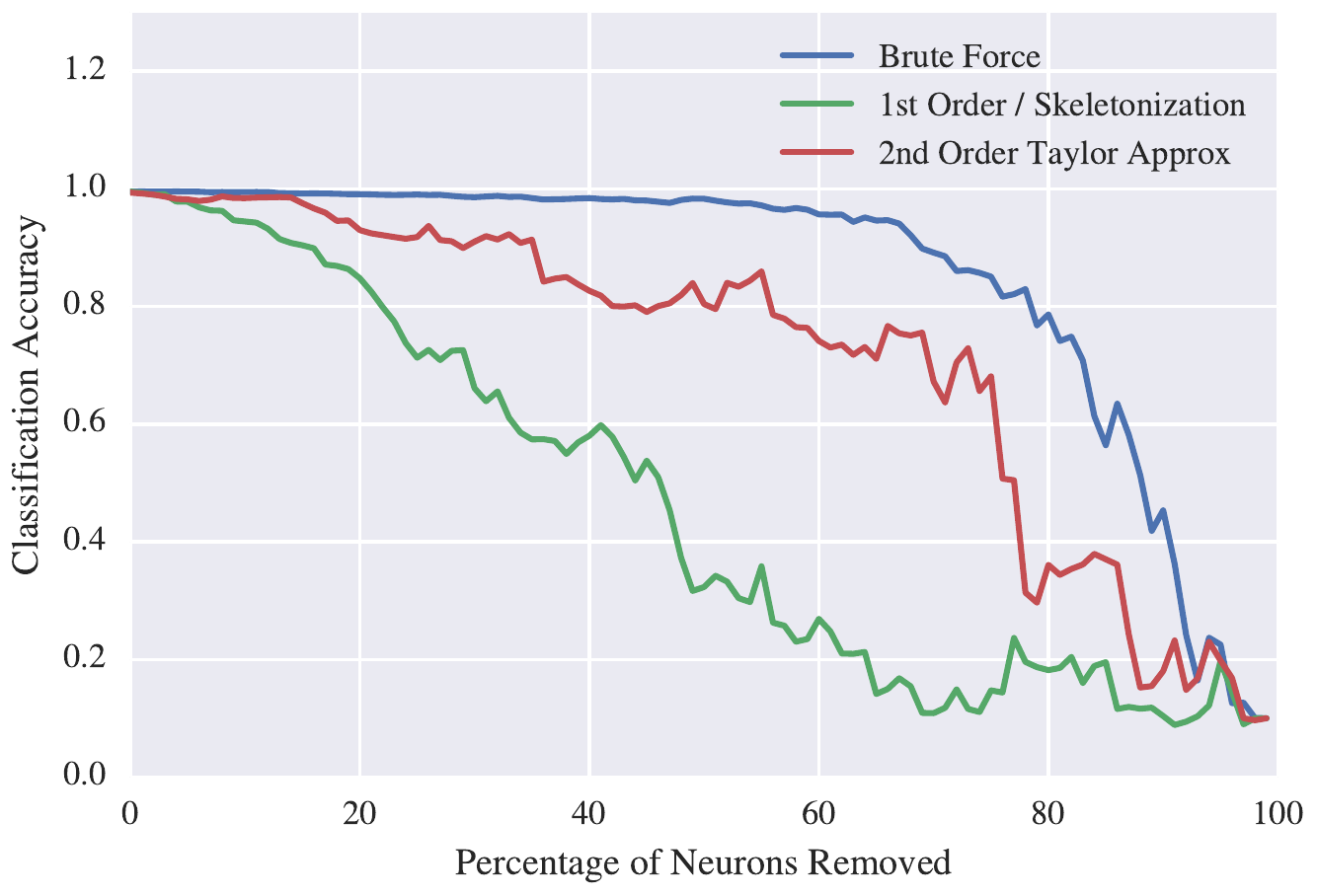}
\includegraphics[width=0.49\linewidth]{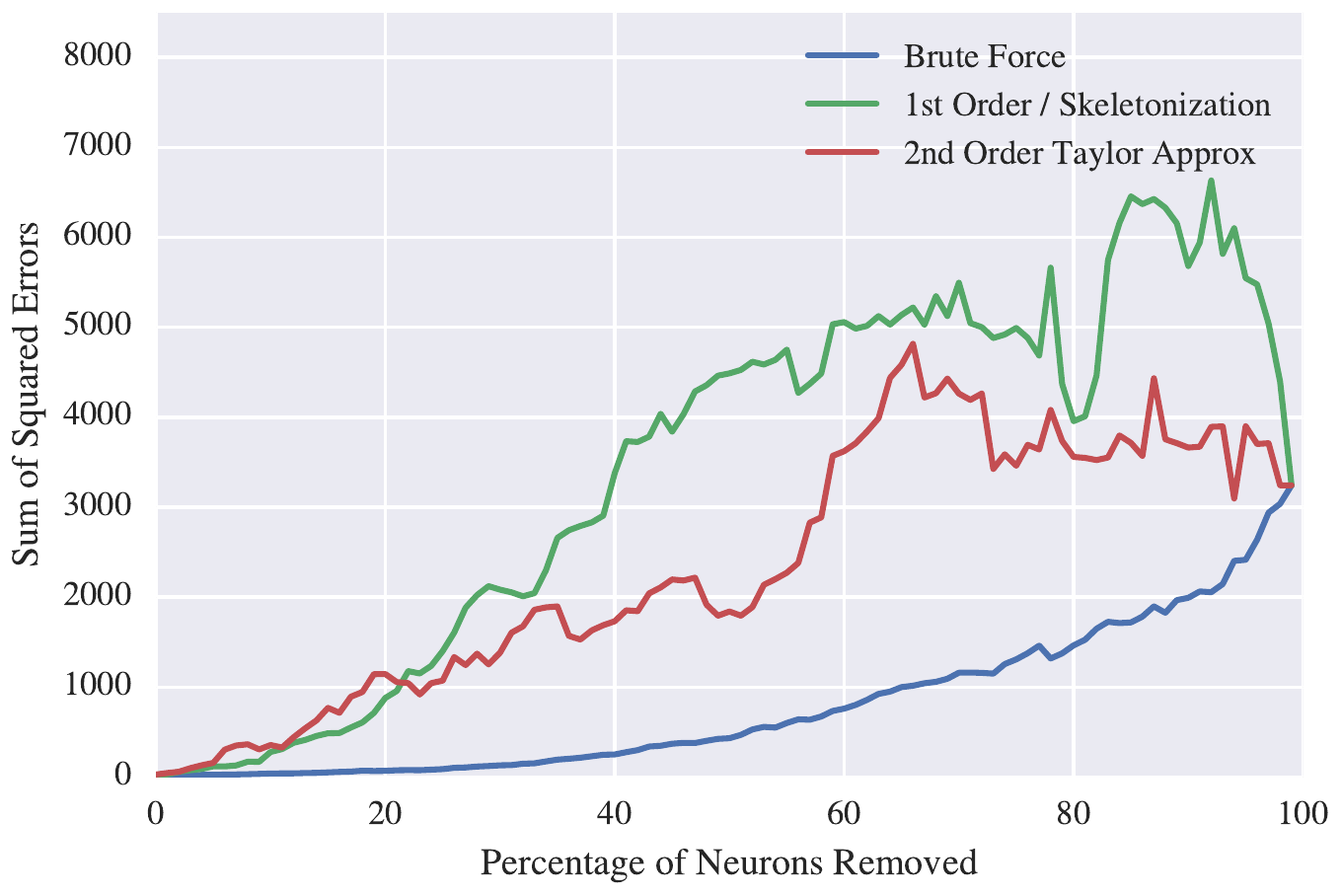}
\includegraphics[width=0.49\linewidth]{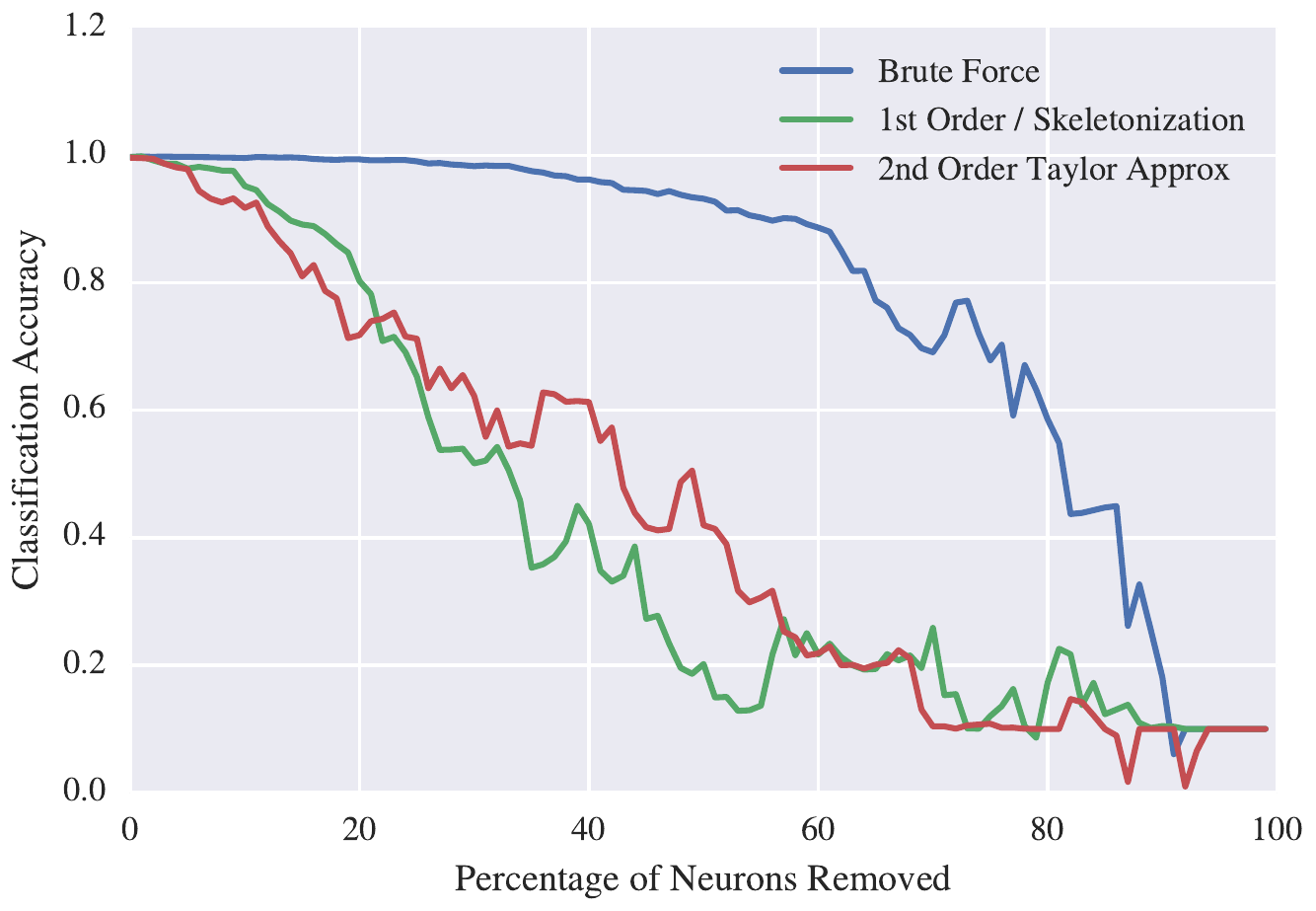}
\caption{Degradation in squared error (left) and classification accuracy (right) after pruning a single-layer network (top) and 2-layer network (bottom) using Algorithm  \ref{algo2} (\textbf{Top Network:} 1 layer, 100 neurons, starting test accuracy: 0.998; \textbf{Bottom Network:} 2 layers, 50 neurons/layer, starting test accuracy: 1.000)}
\label{fig:mnist-re-ranking-single-layer}
\end{figure}

With encouraging observations from our previous experiments, in Figure \ref{fig:mnist-re-ranking-single-layer} we present our results using Algorithm \ref{algo2} (the Iterative Re-ranking Algorithm) on single layer and 2-layer networks trained on the MNIST dataset. There are a few key observations here. Using the brute force ranking criteria, almost 60\% of the neurons in the network can be pruned away without any major loss in performance in the case of a single-layered network. The other noteworthy observation here is that the 2nd order Taylor Series approximation of the error performs consistently better than its 1st order version. It is clear that it becomes harder to remove neurons 1-by-1 with a deeper network, which makes sense because the neurons have more inter-dependencies in that case.  Even with a more complex network, it is possible to remove up to 40\% of the neurons with no major loss in performance which is clearly illustrated by the brute force curve. The more important observation here is the clear potential of an ideal pruning technique and the inconsistency of 1st and 2nd order Taylor Series approximations of the error as ranking criteria.

In order to ensure that our observations were not just a result of idiosyncrasies of the MNIST dataset, we did the same experiment on two toy datasets (Figure \ref{fig:diamond}). These datasets were simple pattern recognition problems where the network's job was to predict whether a given point would lie inside a given shape on the Cartesian plane. The performance of the 2nd-order method was found to be exceptionally good and produced results very close to the brute force method, although it demonstrated a very irregular decline past the tipping point. The 1st-order method, as expected, performed poorly here as well.

\begin{figure}[!ht]
\centering
\includegraphics[width=0.49\linewidth]{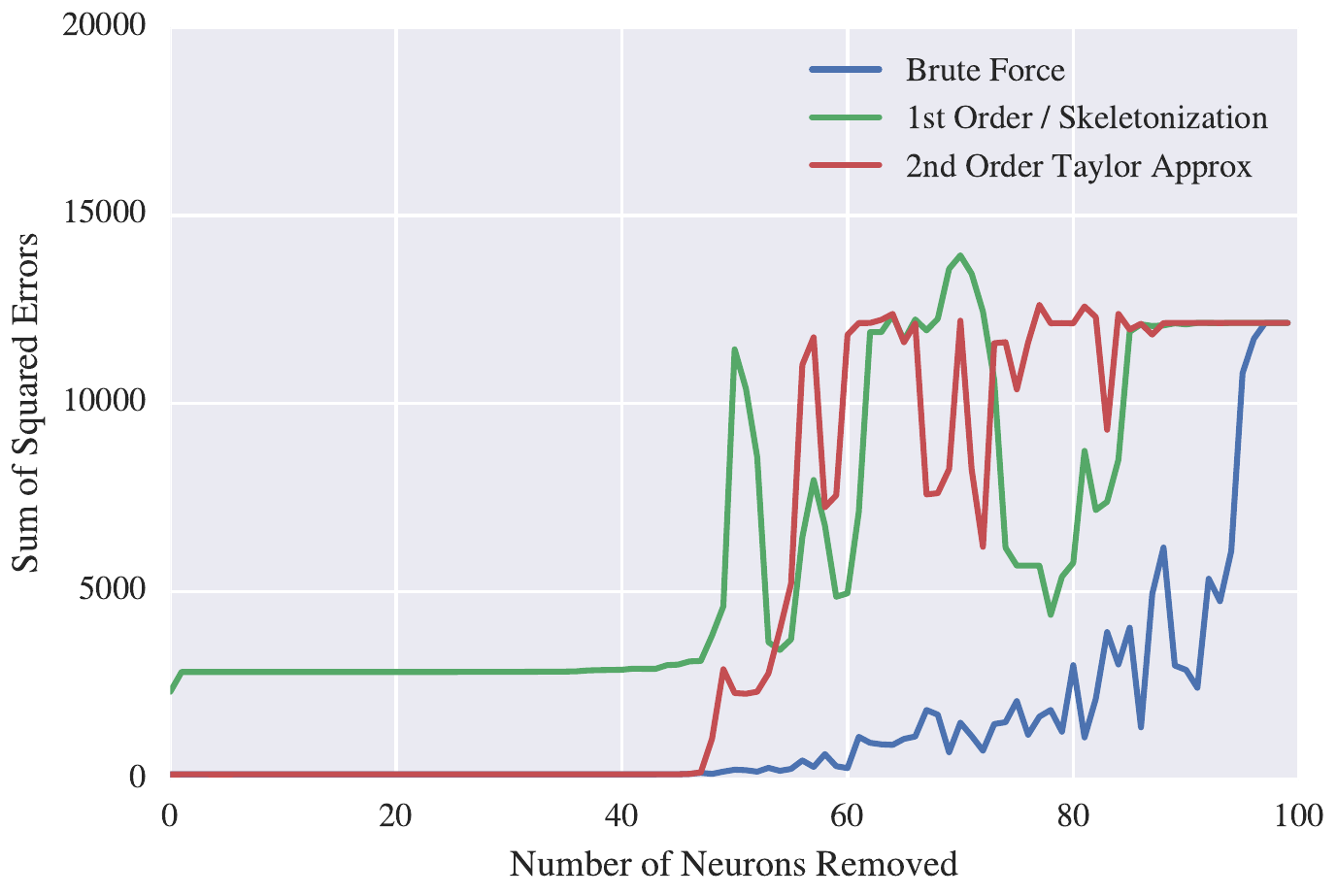}
\includegraphics[width=0.49\linewidth]{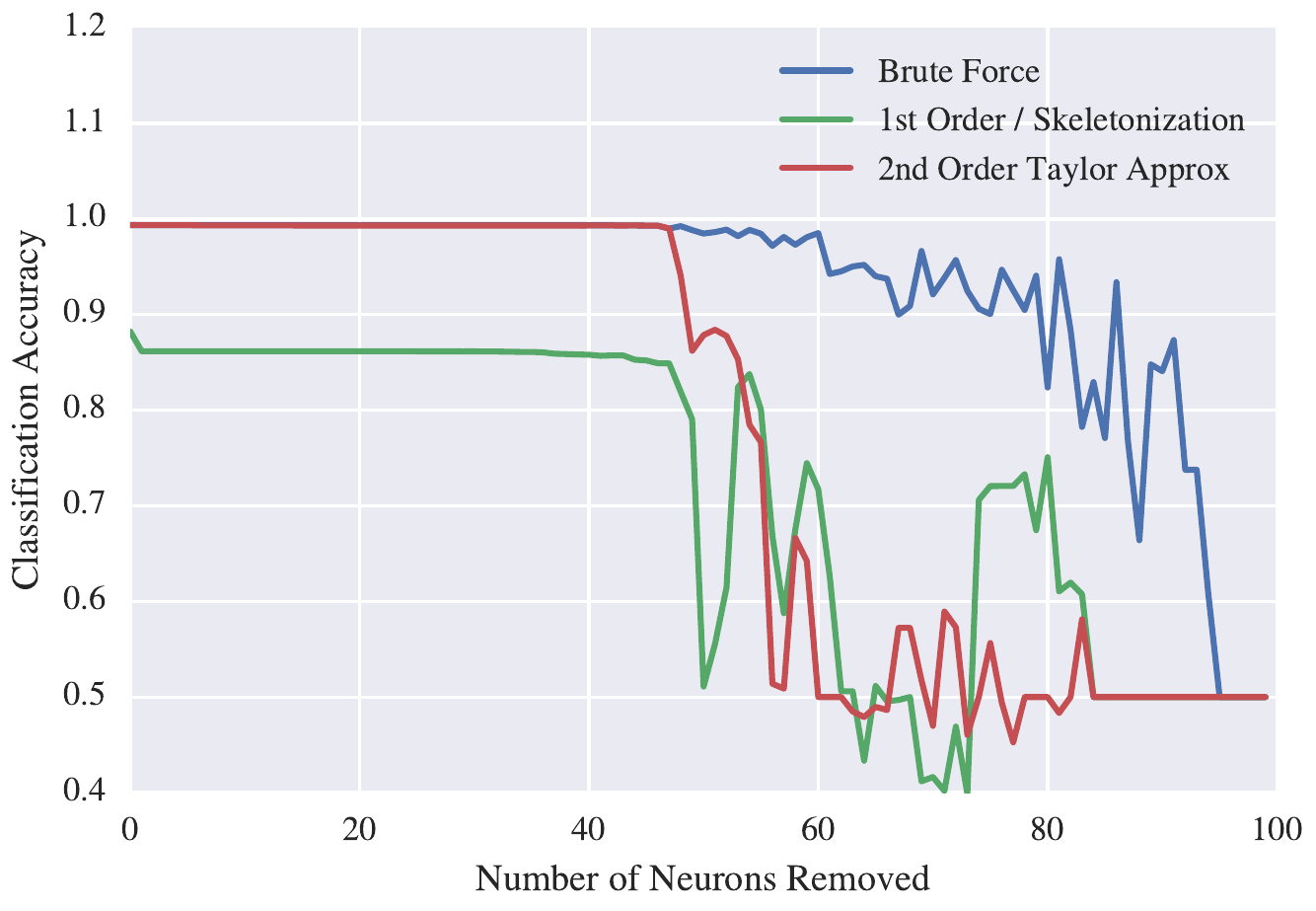}
\includegraphics[width=0.49\linewidth]{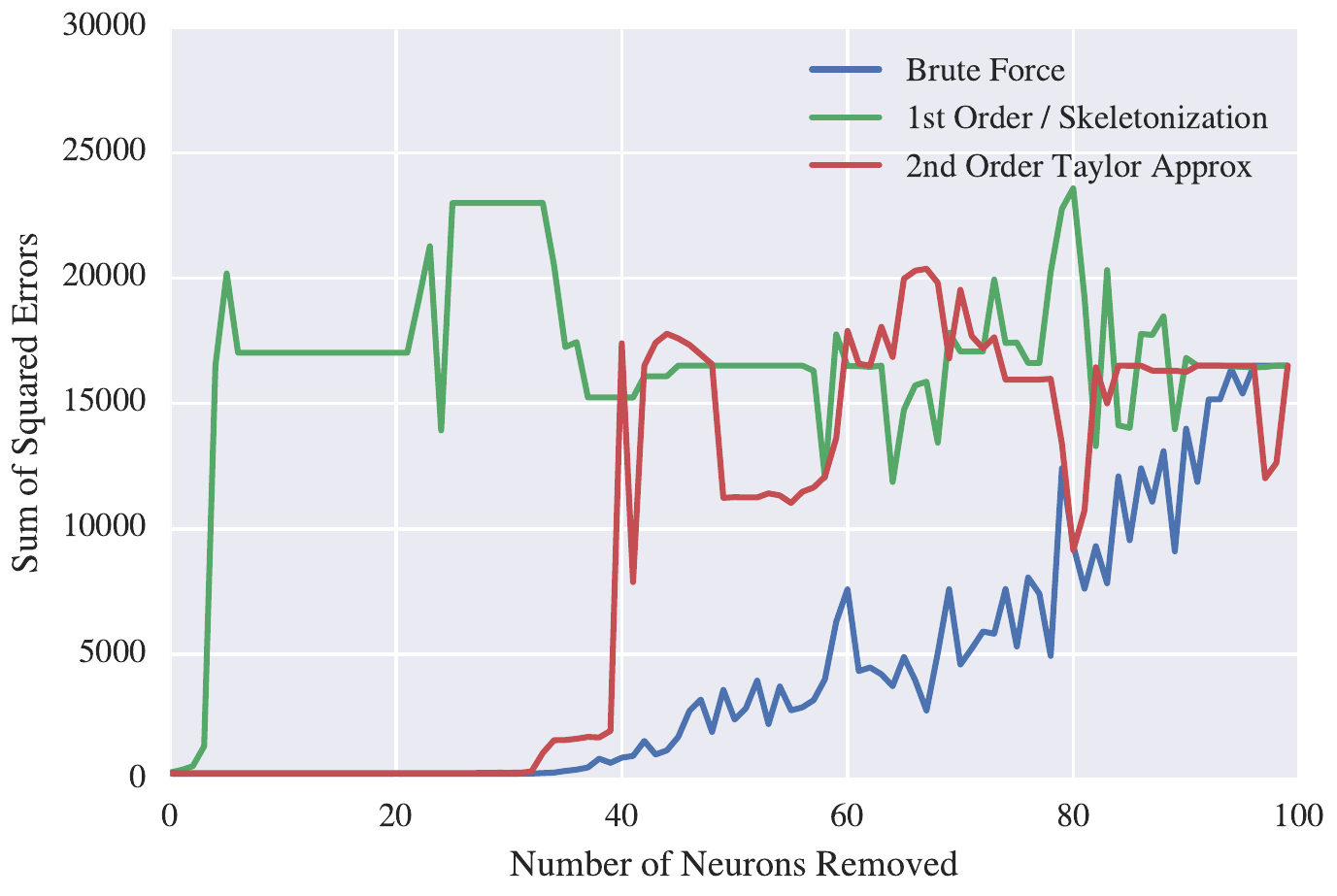}
\includegraphics[width=0.49\linewidth]{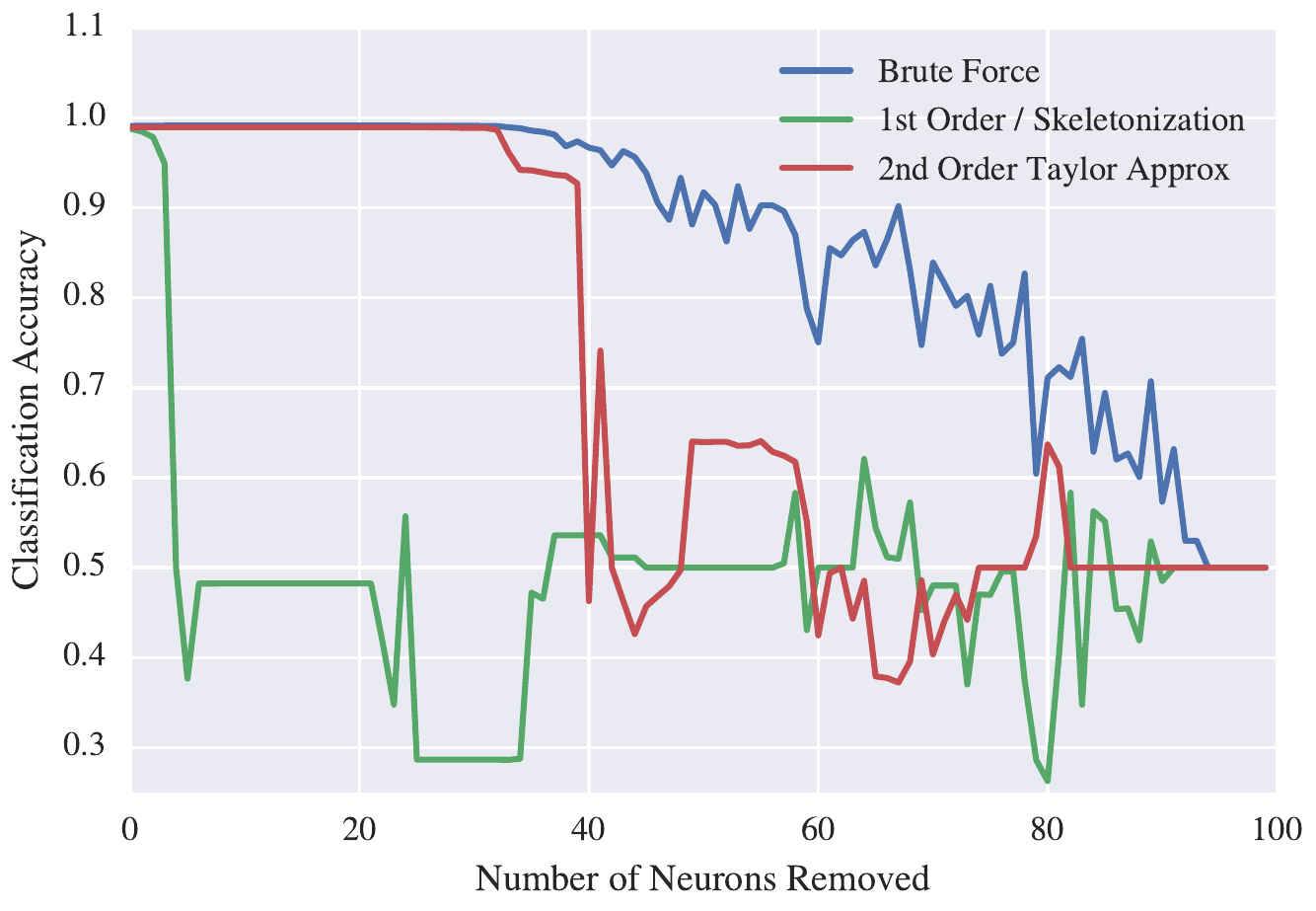}
\caption{Degradation in squared error (left) and classification accuracy (right) after pruning a 2-layer network using Algorithm \ref{algo2} on a toy ``diamond'' shape dataset (top) and a toy ``random shape'' dataset (below); (\textbf{Network:} 2 layers, 50 neurons/layer, 10 outputs, logistic sigmoid activation, starting test accuracy: 0.992(diamond); 0.986(random shape)}
\label{fig:diamond}
\end{figure}

\subsection{Visualization of Error Surface \& Pruning Decisions}
\begin{figure}[!ht]
\centering
\includegraphics[width=\linewidth]{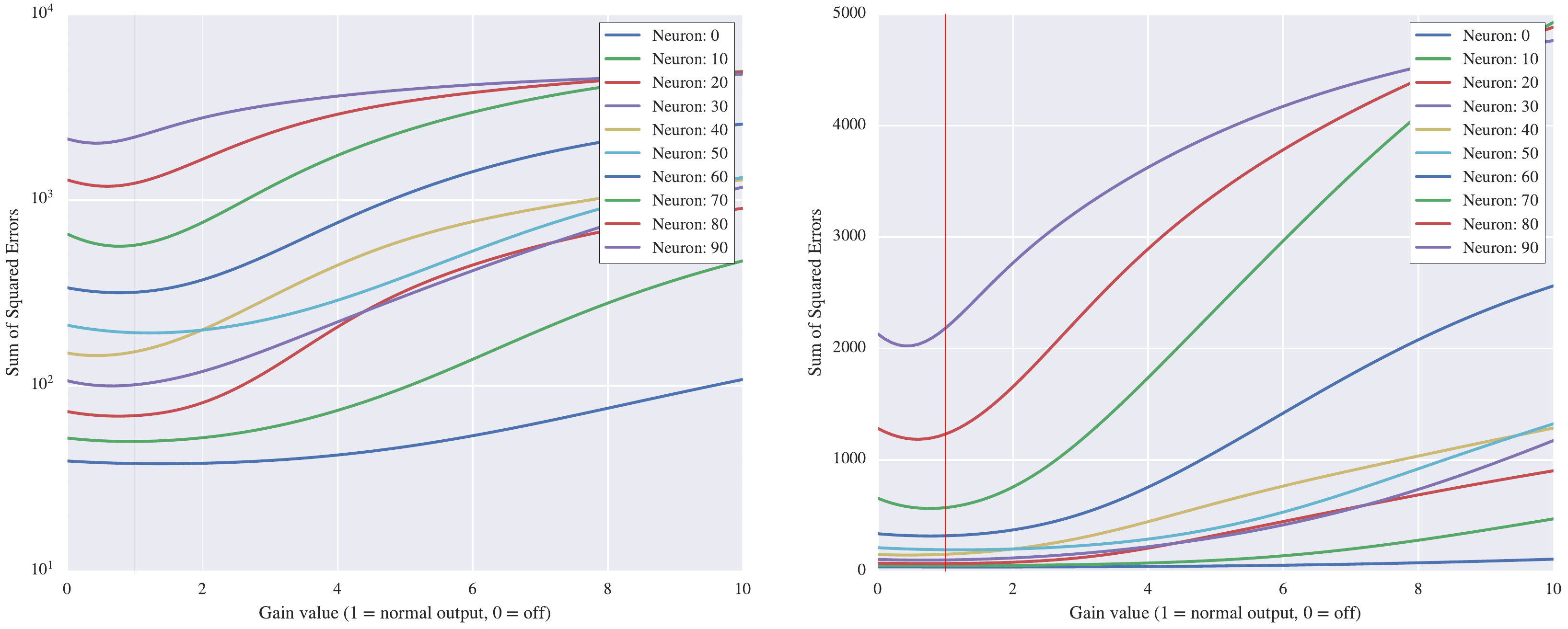}
\includegraphics[width=\linewidth]{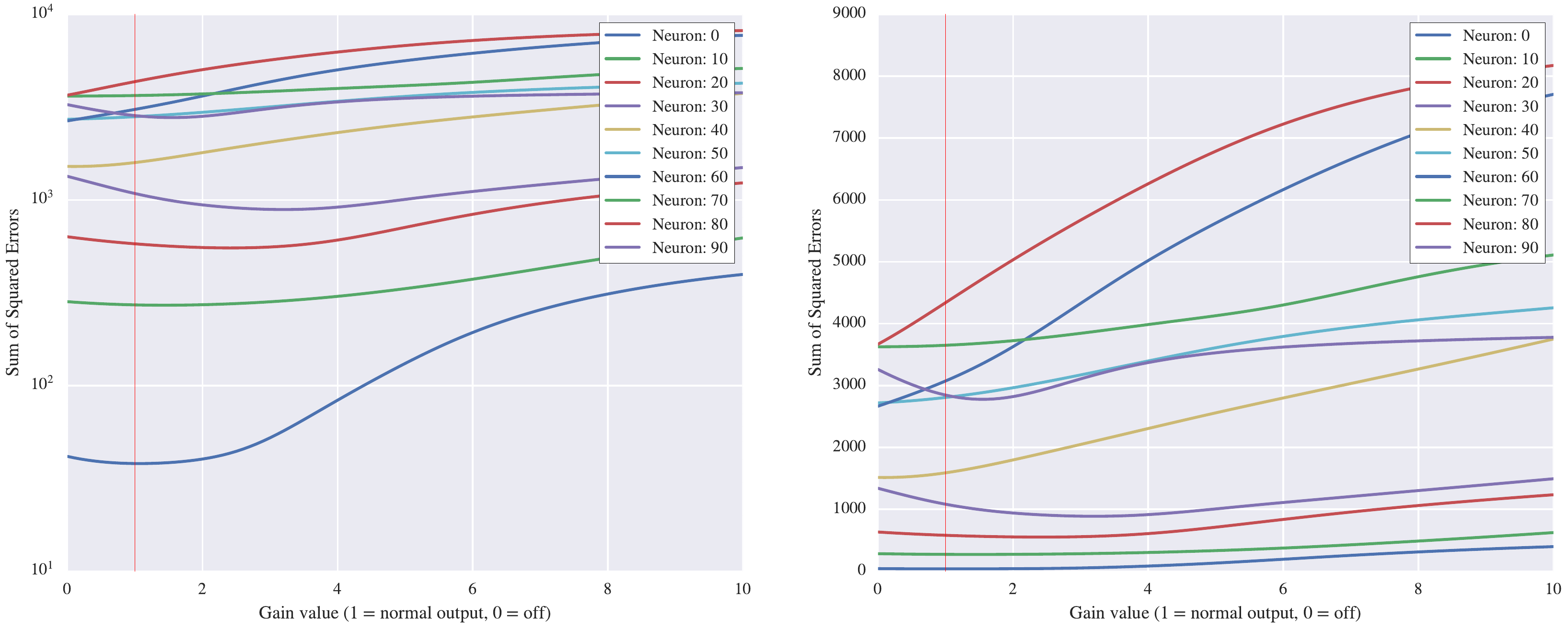}
\includegraphics[width=\linewidth]{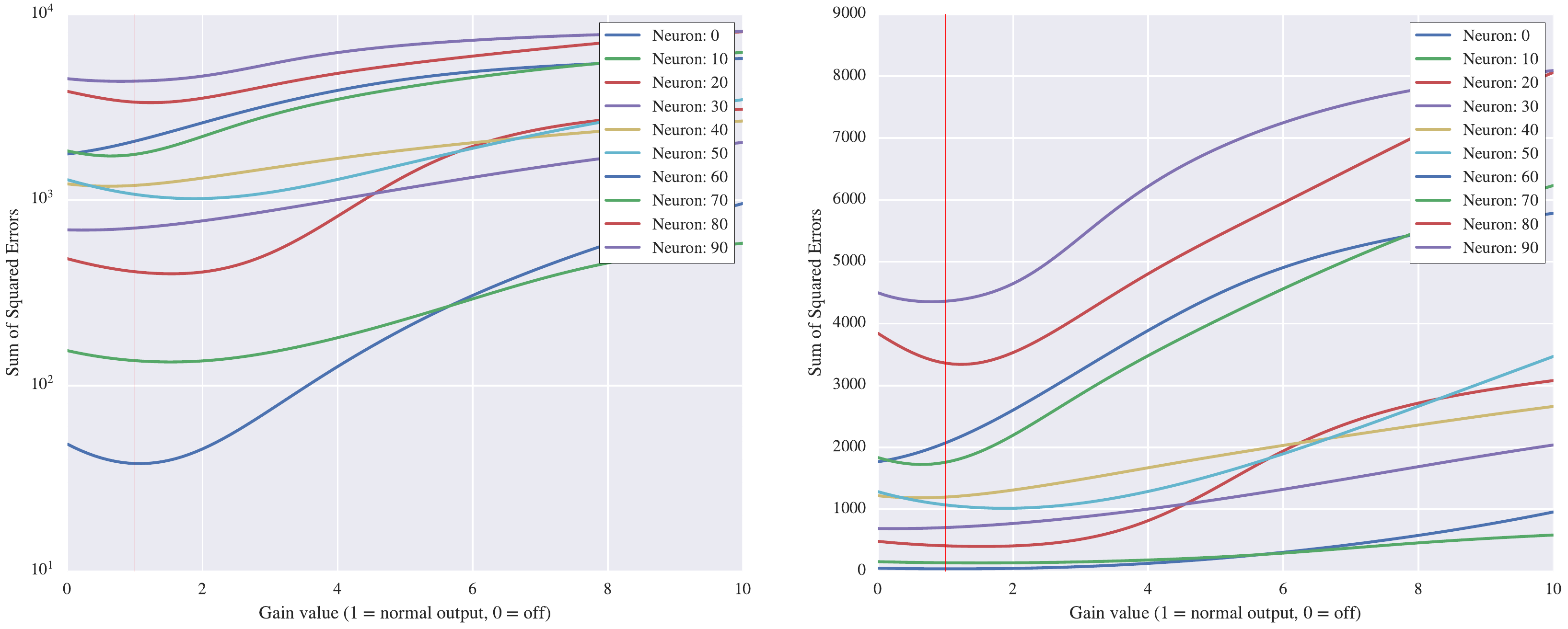}
\caption{Error surface of the network output in log space (left) and real space (right) with respect to each candidate neuron chosen for removal using the brute force, first order and second order criteria}
\label{fig:mnist-single-layer-gt}
\end{figure}

The graphs in Figure \ref{fig:mnist-single-layer-gt} are a visualization of the error surface of the network output with respect to the neurons chosen for removal using each of the 3 ranking criteria, represented in intervals of 10 neurons. Between 0.0 and 1.0, we are graphing the literal effect of turning the neuron off ($\alpha = 0$), and when $\alpha > 1.0$ we are simulating a boosting of the neuron's influence in the network. It can be observed easily that for certain neurons in the network, even doubling, tripling, or quadrupling the scalar output of the neuron has no effect on the overall error of the network, indicating the remarkable degree to which the network has learned to ignore the value of certain parameters. In other cases, we can get a sense of the sensitivity of the network's output to the value of a given neuron when the curve rises steeply after the red 1.0 line. This indicates that the learned value of the parameters emanating from a given neuron are relatively important, and this is why we should ideally see sharper upticks in the curves for the later-removed neurons in the network, that is, when the neurons crucial to the learning representation start to be picked off. As can be easily observed, these functions are too complex to be approximated by straight lines or parabolas, which explains why the popular Taylor Series based methods fail so miserably in our experiments.

\section{Conclusions \& Future Work}\label{sec4}
While this work had never promised proposing a full-fledged pruning algorithm to outperform the existing ones, it had three major goals, all of which it successfully managed to achieve:

\begin{enumerate}
\item Attaining new and deeper insights about learning representations in neural networks.
\item Studying existing pruning techniques to figure out the flaws in the assumptions they make and analyzing these flaws empirically.
\item Using this knowledge to propose a novel paradigm of pruning neural networks and both theoretically and empirically demonstrating its fundamental advantages and correctness over existing techniques.
\end{enumerate}

Through our experiments and exploratory arguments, we were successful in verifying and reinforcing some forgotten observations, were able to answer some previously unanswered questions about learning representations, were able to analyze flaws in existing pruning techniques and finally, were able to propose a novel approach to pruning along with demonstrating its benefits over traditional approaches.

We started our exploratory journey by looking at historical literature in the field of neural network pruning. We discovered that the work done by \cite{mozer1989skeletonization} was not only the first successful attempt at pruning but it also put forth an important insight regarding the unequal distribution of learning representations in trained neural networks, which was largely forgotten over time. We looked at the development of other popular pruning algorithms over the years and made the observation that these algorithms make assumptions which are not necessarily true. We found that some of the most common mistakes these algorithms made were:

\begin{itemize}
\item Ignoring the fact that in a trained neural network, learning is distributed unequally across neurons.
\item Ignoring the fact that neurons in a trained network, irrespective of their overall relevance, rely on each other and hence should not be treated independently, thereby assuming that pruning can be done in a serial fashion.
\item Assuming that the error function with respect to each individual neuron can be approximated with a straight line or more precisely with a parabola. 
\end{itemize}

We proposed a novel approach to neural network pruning as we argued in favor of pruning whole neurons instead of individual weights in Section \ref{sec2}, citing multiple theoretical and logical reasons. We were later able to confirm the feasibility and correctness of this approach empirically through our experiments in Section \ref{sec3}.

One major accomplishment of this work was to answer the previously unanswered question about the relevance of unnecessary neurons in a trained network and the existence of inter-dependencies between relevant and irrelevant neurons. We were able to demonstrate empirically in Section \ref{sec3.1} that these irrelevant neurons do have a role to play in a fully-trained network as the relevant neurons of the network rely on them. Without going into the mathematics of this relationship we were able to demonstrate that the removal of one irrelevant neuron changes the dynamics of the network, following which the contributions of all the remaining neurons need to be reassessed, something that most popular pruning algorithms largely ignore.

Through our experiment on the Example Regression Problem in Section \ref{sec3.2}, we were able to conclusively and empirically verify the original observation made by \cite{mozer1989skeletonization} regarding the unequal and inequitable distribution of learning representations in a trained network. We also saw a strong evidence in support of this during our experiments on visualizing the error surfaces.

Through our experiments in Section \ref{sec3.3}, we were able to demonstrate the flaws in popular pruning techniques which use a linear or quadratic error approximation function. These techniques performed poorly as compared to the simple brute-force pruning method which did not make any faulty assumptions about learning representations or error approximations. We were also able to analyze actual error surfaces which gave us a deeper insight into how these algorithms make bad pruning decisions by approximating complex functions using straight lines and parabolas. We saw how easy was it for these methods to get hopelessly lost after making a bad pruning decision resulting in faults.

Furthermore, we found that the brute-force algorithm does surprisingly well, despite being computationally expensive. We argue that further investigation is warranted to make this algorithm computationally tractable. We have observed that pruning whole neurons from an optimally trained network without major loss in performance is not only possible but also enables compressing networks to 40-70\% of their original size. However expensive, it would be extremely easy to parallelize the brute-force method, or potentially approximate it using a subset of the training data to decide which neurons to prune. This avoids the problem of trying to approximate the importance of a unit and potentially making a mistake. It is also important to note that pruning is only a one-time step in the context of the overall Machine Learning problem and hence its computational complexity can be a good tradeoff for the compression efficiency it brings to the table.

Finally, we encourage the readers of this work to take these results into consideration when making decisions about pruning methods. We also encourage them to leverage this knowledge and insights about learning representations to propose better approximations to the brute force neuron-pruning algorithm. It should be remembered that various heuristics may perform well in practice for reasons which are in fact orthogonal to the accepted justifications given by their proponents.



\bibliography{icml2017_bibliography}
\bibliographystyle{icml2017}  

\newpage
\begin{center}
\appendix{}
\end{center}

\section{Second Derivative Back-Propagation}\label{apd:first}

\begin{figure}[bh!]
\centering
\newcommand{\repSigmoid}{$\sigma(\cdot)$}
\newcommand{\repLinear}{$\sum$}
\newcommand{\repMse}{MSE}
\newcommand{\repFirstSum}{$\Input j 1$}
\newcommand{\repLastSum}{$\Input i 0$}
\newcommand{\repFirstOutput}{\hspace{1.5cm}$\Con j i 0 \!=\! \Weight j i 0 \Out j 1$}
\newcommand{\repLastOutput}{$\Out i 0$}
\newcommand{\repLoss}{$E$}
\def\svgwidth{01\linewidth}
\hspace{-2cm}
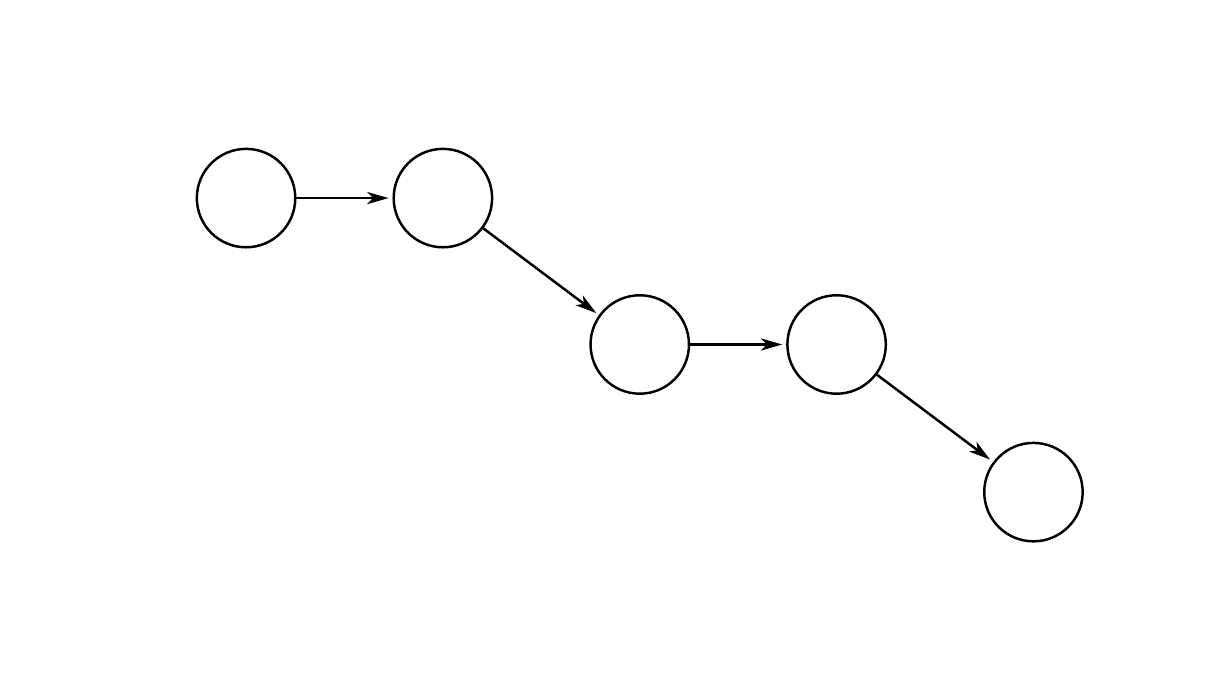
\hspace{-2cm}
\caption{A computational graph of a simple feed-forward network illustrating the naming of different variables, where $\sigma(\cdot)$ is the nonlinearity, MSE is the mean-squared error cost function and $E$ is the overall loss.}
\label{fig:comp_graph}
\end{figure}
Name and network definitions:
\allowbreak
\begin{align}
E &= \frac{1}{2}\sum\limits_i (\Out i 0 - \Target i)^2 &
\Out i m &= \sigma(\Input i m) &
\Input i m &= \sum\limits_j {\Weight j i m}{ \Out j {m + 1}} &
\Con j i m = \Weight j i m \Out j {m+1}
\end{align}
Superscripts represent the index of the layer of the network in question, with 0 representing the output layer. $E$ is the squared-error network cost function. $\Out i m$ is the $i$th output in layer $m$ generated by the activation function $\sigma$, which in this paper is is the standard logistic sigmoid. $\Input i m$ is the weighted sum of inputs to the $i$th neuron in the $m$th layer, and $\Con j i m$ is the contribution of the $j$th neuron in the $m+1$ layer to the input of the $i$th neuron in the $m$th layer. 
\subsection{First and Second Derivatives} 
The first and second derivatives of the cost function with respect to the outputs:
\begin{align}
\pdv{E}{\Out i 0} &= \Out i 0 - \Target i \label{cost_func_derivative}
\end{align}
\begin{align}
\pdv[2]{E}{{\Out i 0}} &= 1\label{cost_func_2nd_derivative}
\end{align}
The first and second derivatives of the sigmoid function in forms depending only on the output:
\begin{align}
\sigma^{\prime}(x) &= \sigma(x)\left(1 - \sigma(x)\right)\label{sigmoid_derivative} 
\\
\sigma^{\prime\prime}(x) &= \sigma^{\prime}(x)\left(1 - 2\sigma(x)\right) \label{sigmoid_2nd_derivative}
\end{align}
The second derivative of the sigmoid is easily derived from the first derivative:
\begin{align}
\sigma^{\prime}(x) &= \sigma(x)\left(1 - \sigma(x)\right)
\\
\sigma^{\prime\prime}(x) &= \dv{}{x}
\underbrace{\sigma(x)}_{f(x)}
\underbrace{\left(1 - \sigma(x)\right)}_{g(x)}
\\
\sigma^{\prime\prime}(x) &= f^{\prime}(x)g(x) + f(x)g^{\prime}(x)
\\
\sigma^{\prime\prime}(x) &= \sigma^{\prime}(x)(1-\sigma(x)) - \sigma(x)\sigma^{\prime}(x)
\\
\sigma^{\prime\prime}(x) &= \sigma^{\prime}(x) - 2\sigma(x)\sigma^{\prime}(x)
\\
\sigma^{\prime\prime}(x) &= \sigma^{\prime}(x)(1 - 2\sigma(x))
\end{align}

And for future convenience: 
\begin{align}
\dv{\Out i m}{\Input i m} &= 
\dv{}{\Input i m}\left({\Out i m} = \sigma(\Input i m)\right) 
\\
&= \left(\Out i m\right)\left(1 - \Out i m\right)
\\
&= \sigma^{\prime}\left(\Input i m\right)
\\
\dv[2]{{\Out i m}}{{\Input i m}} &=
\dv{}{{\Input i m}}\left(\dv{\Out i m}{\Input i m} = \left(\Out i m\right)\left(1 - \Out i m\right)\right)
\\
&= \left(\Out i m\left(1 - \Out i m\right)\right)\left(1 - 2\Out i m\right)
\\
&= \sigma^{\prime\prime}\left(\Input i m\right)
\end{align}

Derivative of the error with respect to the $i$th neuron's input $\Input i 0$ in the output layer:
\begin{align}
\pdv{E}{\Input i 0} &= \pdv{E}{\Out i 0} \pdv{\Out i 0}{\Input i 0} 
\\
&= \underbrace{\left(\Out i 0 - \Target i\right)}_{\text{from} \ (\ref{cost_func_derivative})} \underbrace{\sigma\left(\Input i 0\right)\left(1 - \sigma\left(\Input i 0\right)\right)}
_{\text{from} \ (\ref{sigmoid_derivative})}
\\
&= \left(\Out i 0 - \Target i\right)\left(\Out i 0 \left(1 - \Out i 0\right)\right)
\\
&= \left(\Out i 0 - \Target i\right)\sigma^{\prime}\left(\Input i 0\right)\label{dedx}
\end{align}

Second derivative of the error with respect to the $i$th neuron's input $\Input i 0$ in the output layer:
\begin{align}
\pdv[2]{E}{{\Input i 0}} &= \pdv{}{\Input i 0}
\left(\pdv{E}{\Out i 0}\pdv{\Out i 0}{\Input i 0}\right) 
\\
&= \pdv{E}{\Input i 0}{\Out i 0}
\pdv{\Out i 0}{\Input i 0} + \pdv{E}{\Out i 0}\pdv[2]{\Out i 0}{{\Input i 0}}
\\
&= \pdv{E}{\Input i 0}{\Out i 0}
\underbrace{\left(\Out i 0 \left(1 - \Out i 0\right)\right)}_{{\text{from} \ (\ref{sigmoid_derivative})}} + \underbrace{\left(\Out i 0 - \Target i\right)}_{\text{from} \ (\ref{cost_func_derivative})}\underbrace{\left(\Out i 0 \left(1 - \Out i 0\right)\right)\left(1 - 2\Out i 0\right) }_{\text{from}\ (\ref{sigmoid_2nd_derivative})}
\\
\left(\pdv{E}{\Input i 0}{\Out i 0}\right)
&= \pdv{}{\Input i 0}\pdv{E}{\Out i 0} = \pdv{}{\Input i 0}\underbrace{\left(\Out i 0 - \Target i\right)}_{\text{from} \ (\ref{cost_func_derivative})} = \pdv{\Out i 0}{\Input i 0} = \underbrace{\left(\Out i 0 \left(1 - \Out i 0\right)\right)}_{{\text{from} \ (\ref{sigmoid_derivative})}} 
\\
\pdv[2]{E}{{\Input i 0}}&= \left(\Out i 0 \left(1 - \Out i 0\right)\right)^2 + \left(\Out i 0 - \Target i\right)\left(\Out i 0 \left(1 - \Out i 0\right)\right)\left(1 - 2\Out i 0\right) 
\\
&= \left(\sigma^{\prime}\left(\Input i 0\right)\right)^2 + \left(\Out i 0 - \Target i\right)\sigma^{\prime\prime}\left(\Input i 0\right)\label{d2edx2}
\end{align}

First derivative of the error with respect to a single input contribution $\Con j i 0$ from neuron $j$ to neuron $i$ with weight $\Weight j i 0$ in the output layer:
\begin{align}
\pdv{E}{\Con j i 0} &= 
\pdv{E}{\Out i 0}
\pdv{\Out i 0}{\Input i 0}
\pdv{\Input i 0}{\Con j i 0}
\\
&= \underbrace{\left(\Out i 0 - \Target i \right)}_{\text{from} \ (\ref{cost_func_derivative})} \underbrace{\left(\Out i 0 \left(1 - \Out i 0\right) \right)}_{\text{from} \ (\ref{sigmoid_derivative})} \pdv{\Input i 0}{\Con j i 0} 
\\
\left( \pdv{\Input i m}{\Con j i m}\right) &= \pdv{}{\Con j i m}\left(\Input i m = \sum_j\Weight j i m\Out j {m+1} \right) = \pdv{}{\Con j i m} \left(\Con j i m + k \right) = 1\label{dxdc} 
\\
\pdv{E}{\Con j i 0}&= \left(\Out i 0 - \Target i \right) \left(\Out i 0 \left(1 - \Out i 0\right) \right)
\\
&= \underbrace{\left(\Out i 0 - \Target i \right) \sigma^{\prime}\left(\Input i 0\right)\label{dedc}}
_{\text{from} \ (\ref{dedx})} 
\\
\pdv{E}{\Con j i 0} &= \pdv{E}{\Input i 0}
\end{align}

Second derivative of the error with respect to a single input contribution $\Con j i 0$:
\begin{align}
\pdv[2]{E}{{\Con j i 0}} &=
\pdv{}{\Con j i 0} 
\left(\pdv{E}{\Con j i 0} = 
\underbrace{\left(\Out i 0 - \Target i \right) \sigma^{\prime}\left(\Input i 0\right)}
_{\text{from} \ (\ref{dedc})}
\right)
\\
&=\pdv{}{\Con j i 0}\left(\sigma\left(\Input i 0\right) - \Target i \right) \sigma^{\prime}\left(\Input i 0\right)
\\
&=\pdv{}{\Con j i 0}\left(\sigma\left(\sum\limits_j {\Weight j i m}{ \Out j {m + 1}}\right) - \Target i \right) \sigma^{\prime}\left(\sum\limits_j {\Weight j i m}{ \Out j {m + 1}}\right)
\\
&=\pdv{}{\Con j i 0}\left(\sigma\left(\sum\limits_j {\Con j i 0}\right) - \Target i \right) \sigma^{\prime}\left(\sum\limits_j {\Con j i 0}\right)
\\
&=\pdv{}{\Con j i 0}
\underbrace{\left(\sigma\left({\Con j i 0} + k\right) - \Target i \right)}
_{f\left(\Con j i 0\right)}
\underbrace{\sigma^{\prime}\left({\Con j i 0} + k\right)}
_{g\left(\Con j i 0\right)}
\end{align}

We now make use of the abbreviations $f$ and $g$:
\begin{align}
&=f^{\prime}\left(\Con j i 0\right)g\left(\Con j i 0\right) + f\left(\Con j i 0\right)g^{\prime}\left(\Con j i 0\right)
\\
&=\sigma^{\prime}\left({\Con j i 0} + k\right)\sigma^{\prime}\left({\Con j i 0} + k\right) + 
\left(\sigma\left({\Con j i 0} + k\right) - \Target i \right)\sigma^{\prime\prime}\left({\Con j i 0} + k\right)
\\
&=\sigma^{\prime}\left({\Con j i 0} + k\right)^2 + 
\left(\Out i 0 - \Target i \right)\sigma^{\prime\prime}\left({\Con j i 0} + k\right)
\\
&\left(\Con j i 0 + k = \sum_j{\Con j i 0} = \sum\limits_j {\Weight j i m}{ \Out j {m + 1}} = \Input i 0 \right)
\\
\pdv[2]{E}{{\Con j i 0}}&=
\underbrace{\left(\sigma^{\prime}\left(\Input i 0\right)\right)^2 + 
\left(\Out i 0 - \Target i \right)\sigma^{\prime\prime}\left(\Input i 0\right)}
_{\text{from} \ (\ref{d2edx2})}
\\
\pdv[2]{E}{{\Con j i 0}} &= \pdv[2]{E}{{\Input i 0}}
\end{align}

\subsubsection{Summary Of Output Layer Derivatives}
\begin{align}
&\pdv{E}{\Out i 0} = \Out i 0 - \Target i 
&
\pdv[2]{E}{{\Out i 0}} = 1
\end{align}
\begin{align}
&\pdv{E}{\Input i 0} = \left(\Out i 0 - \Target i\right)\sigma^{\prime}\left(\Input i 0\right)
& 
\pdv[2]{E}{{\Input i 0}} = \left(\sigma^{\prime}\left(\Input i 0\right)\right)^2 + \left(\Out i 0 - \Target i\right)\sigma^{\prime\prime}\left(\Input i 0\right)
\end{align}
\begin{align}
&\pdv{E}{{\Con j i 0}} = \pdv{E}{{\Input i 0}}
&
\pdv[2]{E}{{\Con j i 0}} = \pdv[2]{E}{{\Input i 0}}
\end{align}

\subsubsection{Hidden Layer Derivatives}
The first derivative of the error with respect to a neuron with output $\Out j 1$ in the first hidden layer, summing over all partial derivative contributions from the output layer:
\begin{align}
\pdv{E}{\Out j 1} &= 
\sum_i
\pdv{E}{\Out i 0}
\pdv{\Out i 0}{\Input i 0}
\pdv{\Input i 0}{\Con j i 0}
\pdv{\Con j i 0}{\Out j 1}
= 
\sum_i
\underbrace{\left(\Out i 0 - \Target i\right)\sigma^{\prime}\left(\Input i 0\right)}
_{\text{from} \ (\ref{dedx})}
\Weight j i 0
\\
&\pdv{{\Con j i m}}{{\Out j {m+1}}} = \pdv{}{{\Out j {m+1}}}\left(\Con j i m = \Weight j i m\Out j {m+1}\right) = \Weight j i m\label{dcdo}
\\
\pdv{E}{\Out j 1} &= \sum_i\pdv{E}{\Input i 0}\Weight j i 0
\end{align}
Note that this equation does not depend on the specific form of $\pdv{E}{\Input i 0}$, whether it involves a sigmoid or any other activation function. We can therefore replace the specific indexes with general ones, and use this equation in the future.
\begin{align}
\pdv{E}{\Out j {m+1}} &= \sum_i\pdv{E}{\Input i m}\Weight j i m\label{dedo_general}
\end{align}
The second derivative of the error with respect to a neuron with output $\Out j 1$ in the first hidden layer:
\begin{align}
\pdv[2]{E}{{\Out j 1}} &= 
\pdv{}{\Out j 1}
\pdv{E}{\Out j 1}
\\
&= \pdv{}{\Out j 1}
\sum_i\pdv{E}{\Input i 0}\Weight j i 0
\\
&= \pdv{}{\Out j 1}
\sum_i
\left(\Out i 0 - \Target i\right)\sigma^{\prime}\left(\Input i 0\right)\Weight j i 0
\end{align}

If we now make use of the fact, that 
${\Out i 0} = \sigma\left({\Input i 0}\right) = \sigma\left(\sum_j\left({\Weight j i 0}{\Out j 1}\right)\right)$, we can evaluate the expression further.

\begin{align}
\pdv[2]{E}{{\Out j 1}}
&= \pdv{}{\Out j 1}
\sum_i
\underbrace{\left(\sigma\left(\sum_j{\Weight j i 0}{\Out j 1}\right) - \Target i\right)}
_{f\left(\Out j 1\right)}
\underbrace{\sigma^{\prime}\left(\sum_j{\Weight j i 0}{\Out j 1}\right)\Weight j i 0}
_{g\left(\Out j 1\right)}
\\
&=\sum_i\left(f^{\prime}\left(\Out j 1\right)g\left(\Out j 1\right) + f\left(\Out j 1\right)g^{\prime}\left(\Out j 1\right)\right)
\\
&=\sum_i
\sigma^{\prime}\left(\sum_j{\Weight j i 0}{\Out j 1}\right)\Weight j i 0 \
\sigma^{\prime}\left(\sum_j{\Weight j i 0}{\Out j 1}\right)\Weight j i 0
+ \ldots \\
&\sum_i
\left(\sigma\left(\sum_j{\Weight j i 0}{\Out j 1}\right) - \Target i\right)
\sigma^{\prime\prime}\left(\sum_j{\Weight j i 0}{\Out j 1}\right)\left(\Weight j i 0\right)^2
\\
&=
\sum_i\left(
\left(\sigma^{\prime}\left(\Input i 0\right)\right)^2\left({\Weight j i 0}\right)^2
+ 
\left(\Out i 0 - \Target i\right)
\sigma^{\prime\prime}\left(\Input i 0\right)\left({\Weight j i 0}\right)^2
\right)
\\
&=\sum_i
\underbrace{\left(\left(\sigma^{\prime}\left(\Input i 0\right)\right)^2
+ 
\left(\Out i 0 - \Target i\right)
\sigma^{\prime\prime}\left(\Input i 0\right)\right)}
_{\text{from} \ (\ref{d2edx2})}
\left({\Weight j i 0}\right)^2
\end{align}

Summing up, we obtain the more general expression:
\begin{align}
\pdv[2]{E}{{\Out j 1}} &= 
\sum_i\pdv[2]{E}{{\Input i 0}} \left({\Weight j i 0}\right)^2\label{d2edo2}
\end{align}
Note that the equation in (\ref{d2edo2}) does not depend on the form of $\pdv[2]{E}{{\Input x 0}}$, which means we can replace the specific indexes with general ones:
\begin{align}
\pdv[2]{E}{{\Out j {m+1}}} &= \sum_i
\pdv[2]{E}{{\Input i m}} \left({\Weight j i m}\right)^2\label{de2do2_general}
\end{align} 
At this point we are beginning to see the recursion in the form of the 2nd derivative terms which can be thought of analogously to the first derivative recursion which is central to the back-propagation algorithm. The formulation above which makes specific reference to layer indexes also works in the general case.
\\ 
Consider the $i$th neuron in any layer $m$ with output $\Out i m$ and input $\Input i m$. The first and second derivatives of the error $E$ with respect to this neuron's \textit{input} are: 
\begin{align}
\pdv{E}{\Input i m} &= 
\pdv{E}{\Out i m}
\pdv{\Out i m}{\Input i m}\label{dedx_general}
\end{align}
\begin{align}
\pdv[2]{E}{{\Input i m}} &= 
\pdv{}{{\Input i m}}
\pdv{E}{{\Input i m}} 
\\
&= \pdv{}{\Input i m}
\left(
\pdv{E}{\Out i m}
\pdv{\Out i m}{\Input i m}
\right)
\\
&= \pdv{E}{\Input i m}{\Out i m}
\pdv{\Out i m}{\Input i m}
+
\pdv{E}{\Out i m}\pdv[2]{{\Out i m}}{{\Input i m}}
\\
&=\pdv{}{{\Out i m}}
\left(\pdv{E}{\Input i m} = \pdv{E}{{\Out i m}}\pdv{{\Out i m}}{{\Input i m}}\right)
\pdv{{\Out i m}}{{\Input i m}}
+
\pdv{E}{\Out i m}\sigma^{\prime\prime}\left(\Input i m\right)
\\
&=\pdv[2]{E}{{\Out i m}}
\left
(\pdv{{\Out i m}}{{\Input i m}}
\pdv{{\Out i m}}{{\Input i m}}
\right)
+
\pdv{E}{\Out i m}\sigma^{\prime\prime}\left(\Input i m\right)
\\
\pdv[2]{E}{{\Input i m}} &= 
\pdv[2]{E}{{\Out i m}} \left(\sigma^{\prime}\left({\Input i m}\right)\right)^2
+
\pdv{E}{{\Out i m}}\sigma^{\prime\prime}\left(\Input i m\right)
\end{align}
Note the form of this equation is the general form of what was derived for the output layer in (\ref{d2edx2}). Both of the above first and second terms are easily computable and can be stored as we propagate back from the output of the network to the input. With respect to the output layer, the first and second derivative terms have already been derived above. In the case of the $m + 1$ hidden layer during back propagation, there is a summation of terms calculated in the $m$th layer. For the first derivative, we have this from (\ref{dedo_general}).
\begin{align}
\pdv{E}{\Out j {m+1}} &= \sum_i\pdv{E}{\Input i m}\Weight j i m
\end{align}
And the second derivative for the $j$th neuron in the $m+1$ layer:
\begin{align}
\pdv[2]{E}{{\Input j {m+1}}} &= 
\pdv[2]{E}{{\Out j {m+1}}}
\left(\sigma^{\prime}\left({\Input j {m+1}}\right)\right)^2
+
\pdv{E}{{\Out j {m+1}}}\sigma^{\prime\prime}\left(\Input j {m+1}\right)
\end{align}
We can replace both derivative terms with the forms which depend on the previous layer:
\begin{align}
\pdv[2]{E}{{\Input j {m+1}}} &= 
\underbrace{\sum_i\pdv[2]{E}{{\Input i 0}} \left({\Weight j i 0}\right)^2}
_{\text{from} \ (\ref{de2do2_general})}
\left(\sigma^{\prime}\left({\Input j {m+1}}\right)\right)^2
+
\underbrace{\sum_i\pdv{E}{\Input i m}\Weight j i m}
_{\text{from} \ (\ref{dedo_general})}
\sigma^{\prime\prime}\left(\Input j {m+1}\right)
\end{align}
And this horrible mouthful of an equation gives you a general form for any neuron in the $j$th position of the $m+1$ layer. Taking very careful note of the indexes, this can be more or less translated painlessly to code. You are welcome, world.

\subsubsection{Summary Of Hidden Layer Derivatives}
\begin{align}
\pdv{E}{\Out j {m+1}} &= \sum_i\pdv{E}{\Input i m}\Weight j i m &
\pdv[2]{E}{{\Out j {m+1}}} &= 
\sum_i\pdv[2]{E}{{\Input i m}} \left({\Weight j i m}\right)^2
\end{align}
\begin{align}
\pdv{E}{\Input i m} &= 
\pdv{E}{\Out i m}
\pdv{\Out i m}{\Input i m} \\
\pdv[2]{E}{{\Input j {m+1}}} &= 
\pdv[2]{E}{{\Out j {m+1}}}
\left(\sigma^{\prime}\left({\Input j {m+1}}\right)\right)^2
+
\pdv{E}{{\Out j {m+1}}}\sigma^{\prime\prime}\left(\Input j {m+1}\right)
\end{align}

\section{Additional Experimental Results}\label{apd:second}

\subsection{Example Regression Problem}

\begin{figure}[!ht]
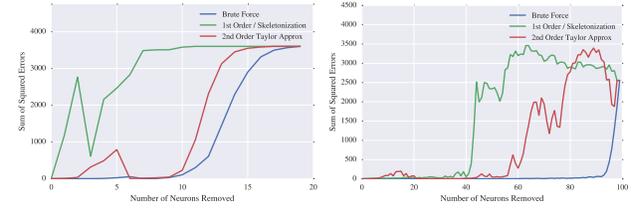

\centering
\includegraphics[width=0.49\linewidth]{cos-small.pdf}
\includegraphics[width=0.49\linewidth]{cos-big.pdf}
\caption{Degradation in squared error after pruning a two-layer network trained to compute the cosine function (\textbf{Left Network:} 2 layers, 10 neurons each, 1 output, logistic sigmoid activation, starting test accuracy: 0.9999993, \textbf{Right Network:} 2 layers, 50 neurons each, 1 output, logistic sigmoid activation, starting test accuracy: 0.9999996)}
\label{fig:cosine-double-layer}
\end{figure}

\subsection{Results on MNIST Dataset}
For all the results presented in this section, the MNIST database of Handwritten Digits by \cite{lecun-mnisthandwrittendigit-2010} was used. It is worth noting that due to the time taken by the brute force algorithm we rather used a 5000 image subset of the MNIST database in which we have normalized the pixel values between 0 and 1.0, and compressed the image sizes to 20x20 images rather than 28x28, so the starting test accuracy reported here appears higher than those reported by LeCun et al. We do not believe that this affects the interpretation of the presented results because the basic learning problem does not change with a larger dataset or input dimension.

\subsection{Pruning a 1-Layer Network}
The network architecture in this case consisted of 1 layer, 100 neurons, 10 outputs, logistic sigmoid activations, and a starting test accuracy of 0.998.

\subsubsection{Single Overall Ranking Algorithm}
\begin{figure}[!ht]
\centering
\includegraphics[width=0.49\linewidth]{mnist-acc99-single-pass-method.pdf}
\includegraphics[width=0.49\linewidth]{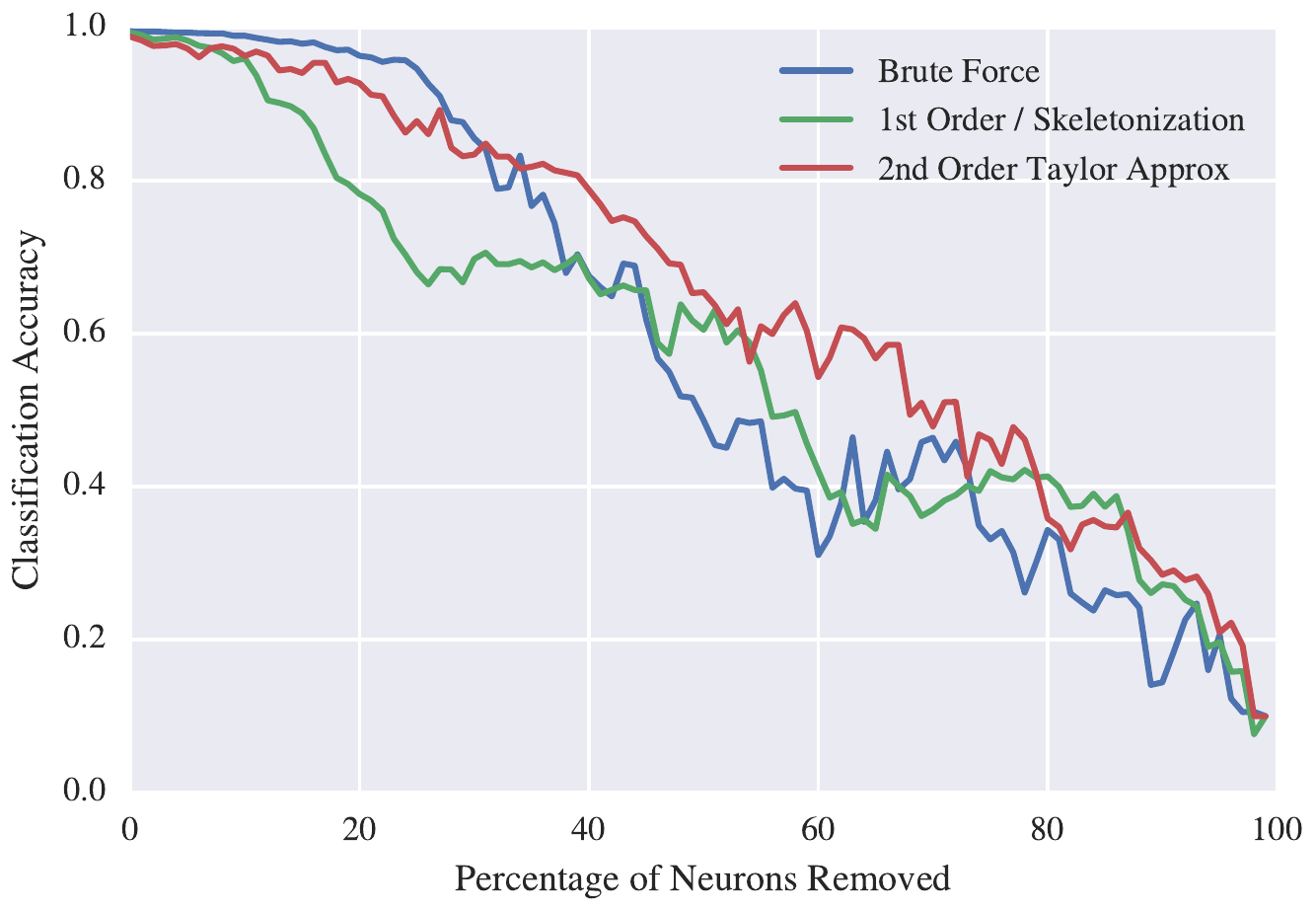}
\caption{Degradation in squared error (left) and classification accuracy (right) after pruning a single-layer network using The Single Overall Ranking algorithm (\textbf{Network:} 1 layer, 100 neurons, 10 outputs, logistic sigmoid activation, starting test accuracy: 0.998)}
\label{fig:mnist-single-ranking-single-layer}
\end{figure}

We first present the results for a single-layer neural network in Figure \ref{fig:mnist-single-ranking-single-layer}, using the Single Overall algorithm (Algorithm \ref{algo1}) as proposed in Section \ref{sec2}. 

\subsubsection{Iterative Re-Ranking Algorithm}
\begin{figure}[!ht]
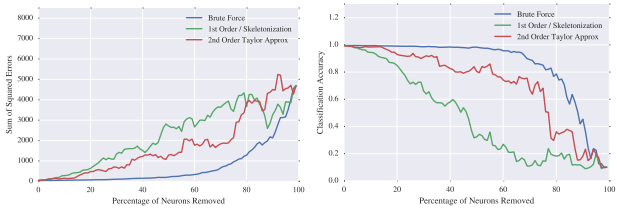

\centering
\includegraphics[width=0.49\linewidth]{mnist-acc99-iterative-rerank-method.pdf}
\includegraphics[width=0.49\linewidth]{mnist-acc99-iterative-rerank-accuracy.pdf}
\caption{Degradation in squared error (left) and classification accuracy (right) after pruning a single-layer network the iterative re-ranking algorithm (\textbf{Network:} 1 layer, 100 neurons, 10 outputs, logistic sigmoid activation, starting test accuracy: 0.998)}
\label{fig:mnist-re-ranking-single-layer}
\end{figure}

\subsubsection{Visualization of Error Surface \& Pruning Decisions}

These graphs are a visualization of the error surface of the network output with respect to the neurons chosen for removal using each of the 3 ranking criteria, represented in intervals of 10 neurons. In each graph, the error surface of the network output is displayed in log space (left) and in real space (right) with respect to each candidate neuron chosen for removal. We create these plots during the pruning exercise by picking a neuron to switch off, and then multiplying its output by a scalar gain value $\alpha$ which is adjusted from 0.0 to 10.0 with a step size of 0.001. When the value of $\alpha$ is 1.0, this represents the unperturbed neuron output learned during training. Between 0.0 and 1.0, we are graphing the literal effect of turning the neuron off ($\alpha = 0$), and when $\alpha > 1.0$ we are simulating a boosting of the neuron's influence in the network, i.e. inflating the value of its outgoing weight parameters. 

We graph the effect of boosting the neuron's output to demonstrate that for certain neurons in the network, even doubling, tripling, or quadrupling the scalar output of the neuron has no effect on the overall error of the network, indicating the remarkable degree to which the network has learned to ignore the value of certain parameters. In other cases, we can get a sense of the sensitivity of the network's output to the value of a given neuron when the curve rises steeply after the red 1.0 line. This indicates that the learned value of the parameters emanating from a given neuron are relatively important, and this is why we should ideally see sharper upticks in the curves for the later-removed neurons in the network, that is, when the neurons crucial to the learning representation start to be picked off. Some very interesting observations can be made in each of these graphs. 

Remember that lower is better in terms of the height of the curve and minimal (or negative) horizontal change between the vertical red line at 1.0 (neuron \textit{on}, $\alpha = 1.0$) and 0.0 (neuron \textit{off}, $\alpha = 0.0$) is indicative of a good candidate neuron to prune, i.e. there will be minimal effect on the network output when the neuron is removed. 

\subsubsection{Visualization of brute force Pruning Decisions}
\begin{figure}[!ht]
\centering
\includegraphics[width=\linewidth]{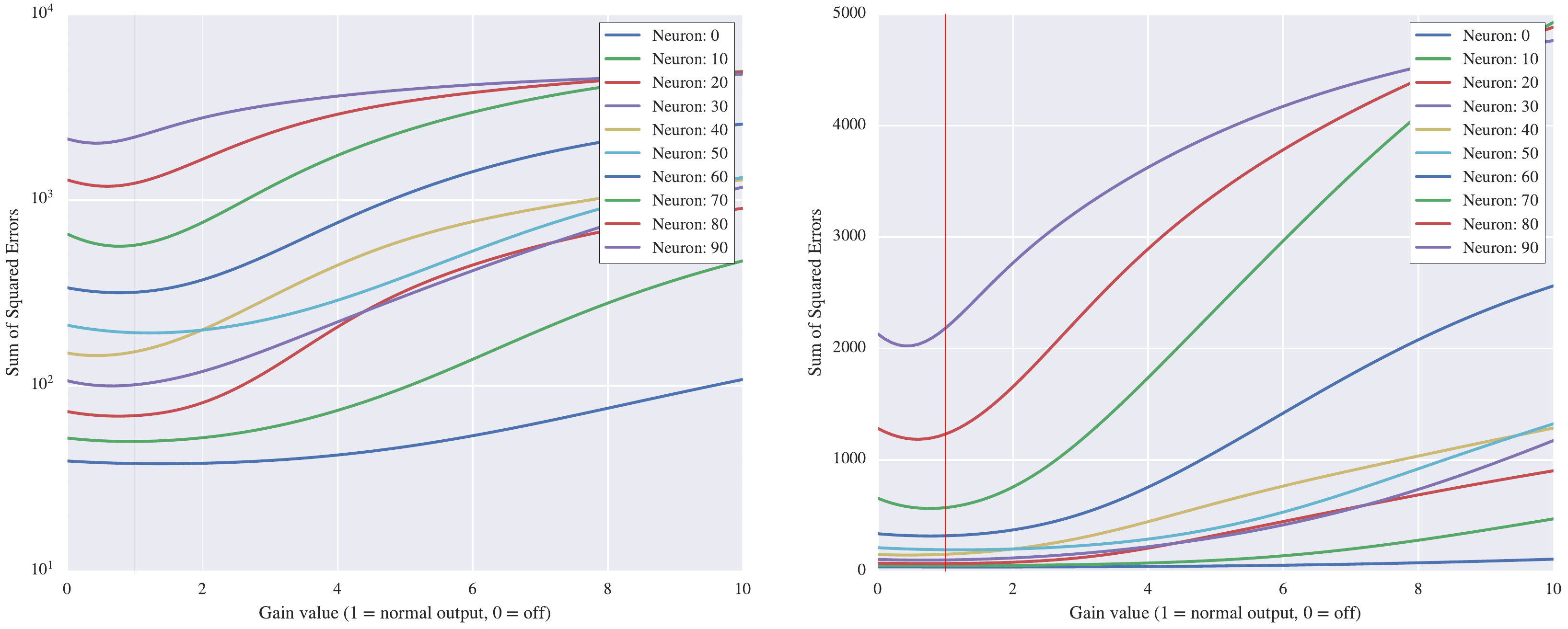}
\caption{Error surface of the network output in log space (left) and real space (right) with respect to each candidate neuron chosen for removal using the brute force criterion; (\textbf{Network:} 1 layer, 100 neurons, 10 outputs, logistic sigmoid activation, starting test accuracy: 0.998)}
\label{fig:mnist-single-layer-gt}
\end{figure}

In Figure \ref{fig:mnist-gt-single-layer}, we notice how low to the floor and flat most of the curves are. It's only until the 90th removed neuron that we see a higher curve with a more convex shape (clearly a more sensitive, influential piece of the network). 

\subsubsection{Visualization of 1st Order Approximation Pruning Decisions}
\begin{figure}[!ht]
\centering
\includegraphics[width=\linewidth]{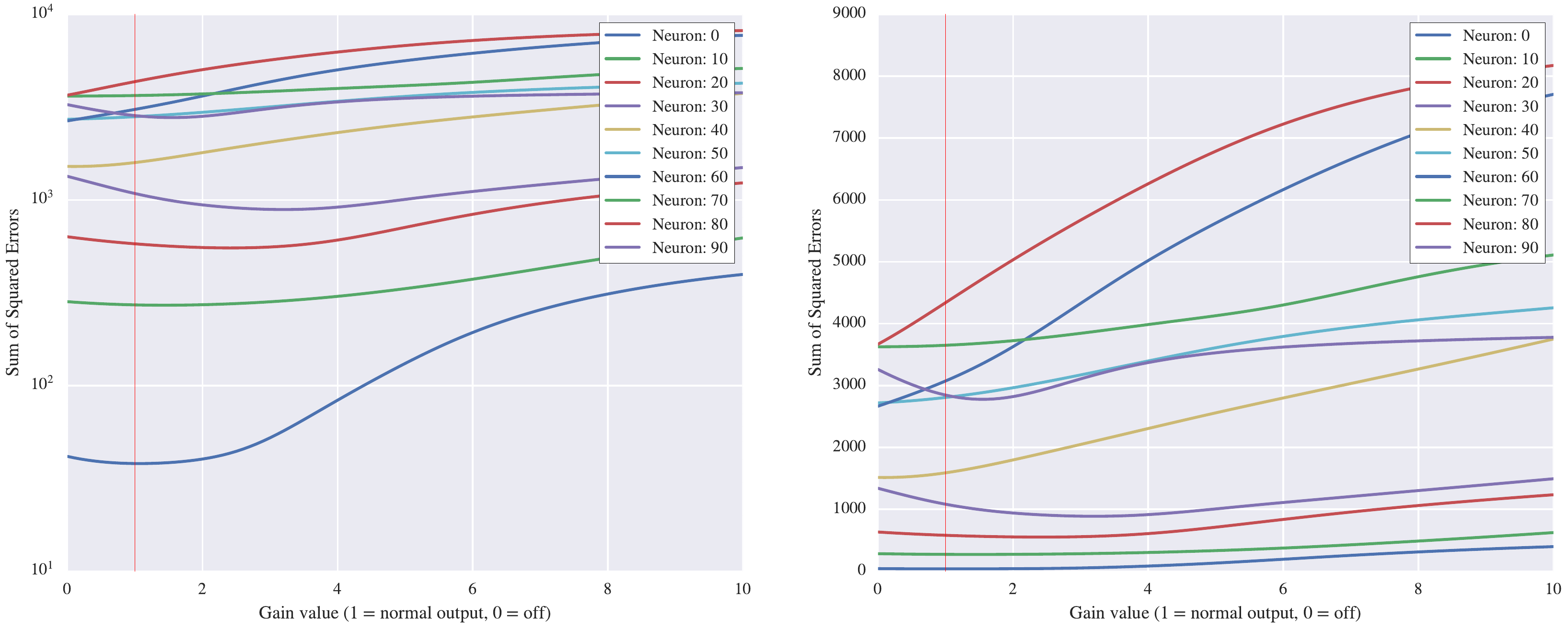}
\caption{Error surface of the network output in log space (left) and real space (right) with respect to each candidate neuron chosen for removal using the 1st order Taylor Series error approximation criterion; (\textbf{Network:} 1 layer, 100 neurons, 10 outputs, logistic sigmoid activation, starting test accuracy: 0.998)}
\label{fig:mnist-single-layer-g1}
\end{figure}

It can be seen in Figure \ref{fig:mnist-single-layer-g1} that most choices seem to have flat or negatively sloped curves, indicating that the first order approximation seems to be pretty good, but examining the brute force choices shows they could be better. 

\subsubsection{Visualization of 2nd Order Approximation Pruning Decisions}
\begin{figure}[!ht]
\centering
\includegraphics[width=\linewidth]{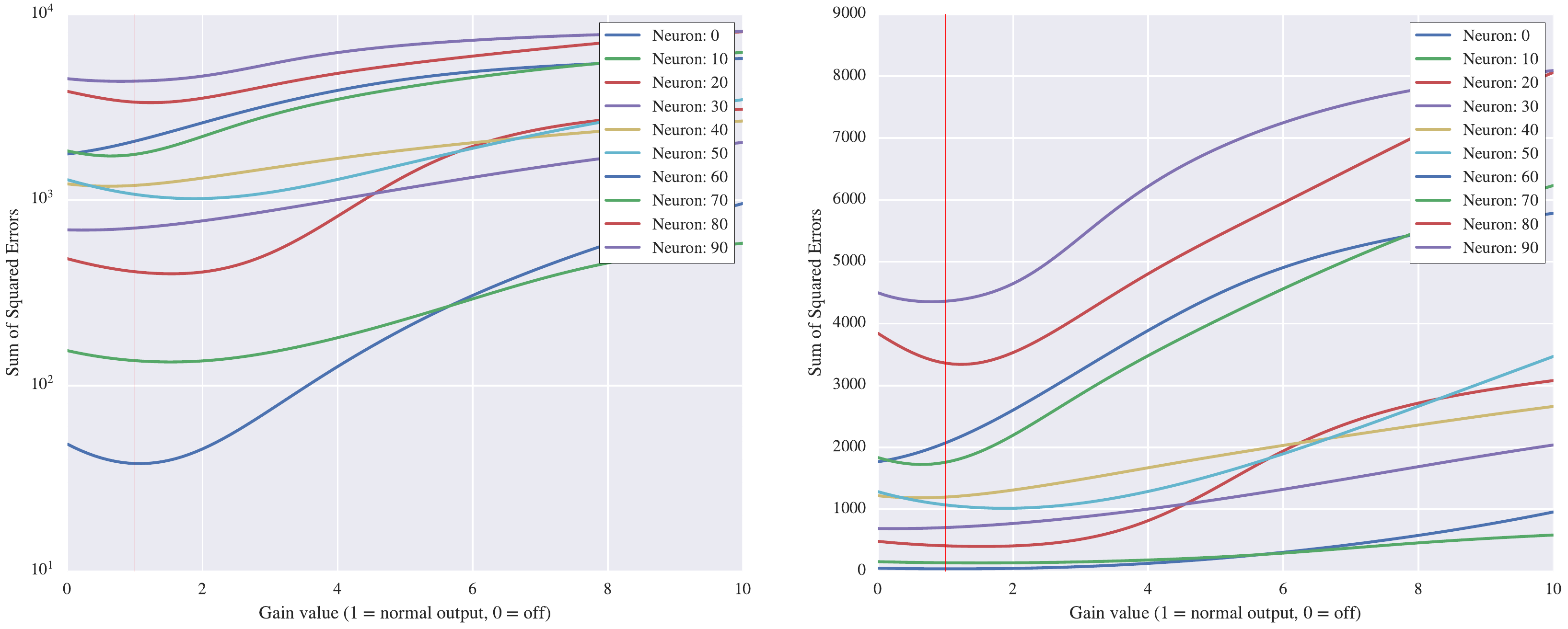}
\caption{Error surface of the network output in log space (left) and real space (right) with respect to each candidate neuron chosen for removal using the 2nd order Taylor Series error approximation criterion; (\textbf{Network:} 1 layer, 100 neurons, 10 outputs, logistic sigmoid activation, starting test accuracy: 0.998)}
\label{fig:mnist-single-layer-g2}
\end{figure}

The method in Figure \ref{fig:mnist-single-layer-g2} looks similar to the brute force method choices, though clearly not as good (they're more spread out). Notice the difference in convexity between the 2nd and 1st order method choices. It's clear that the first order method is fitting a line and the 2nd order method is fitting a parabola in their approximation. 

\subsection{Pruning A 2-Layer Network}
The network architecture in this case consisted of 2 layers, 50 neurons per layer, 10 outputs, logistic sigmoid activation, and a starting test accuracy of 1.000.

\subsubsection{Single Overall Ranking Algorithm}
\begin{figure}[!ht]
\centering
\includegraphics[width=0.49\linewidth]{mnist-deep-single-pass-method.pdf}
\includegraphics[width=0.49\linewidth]{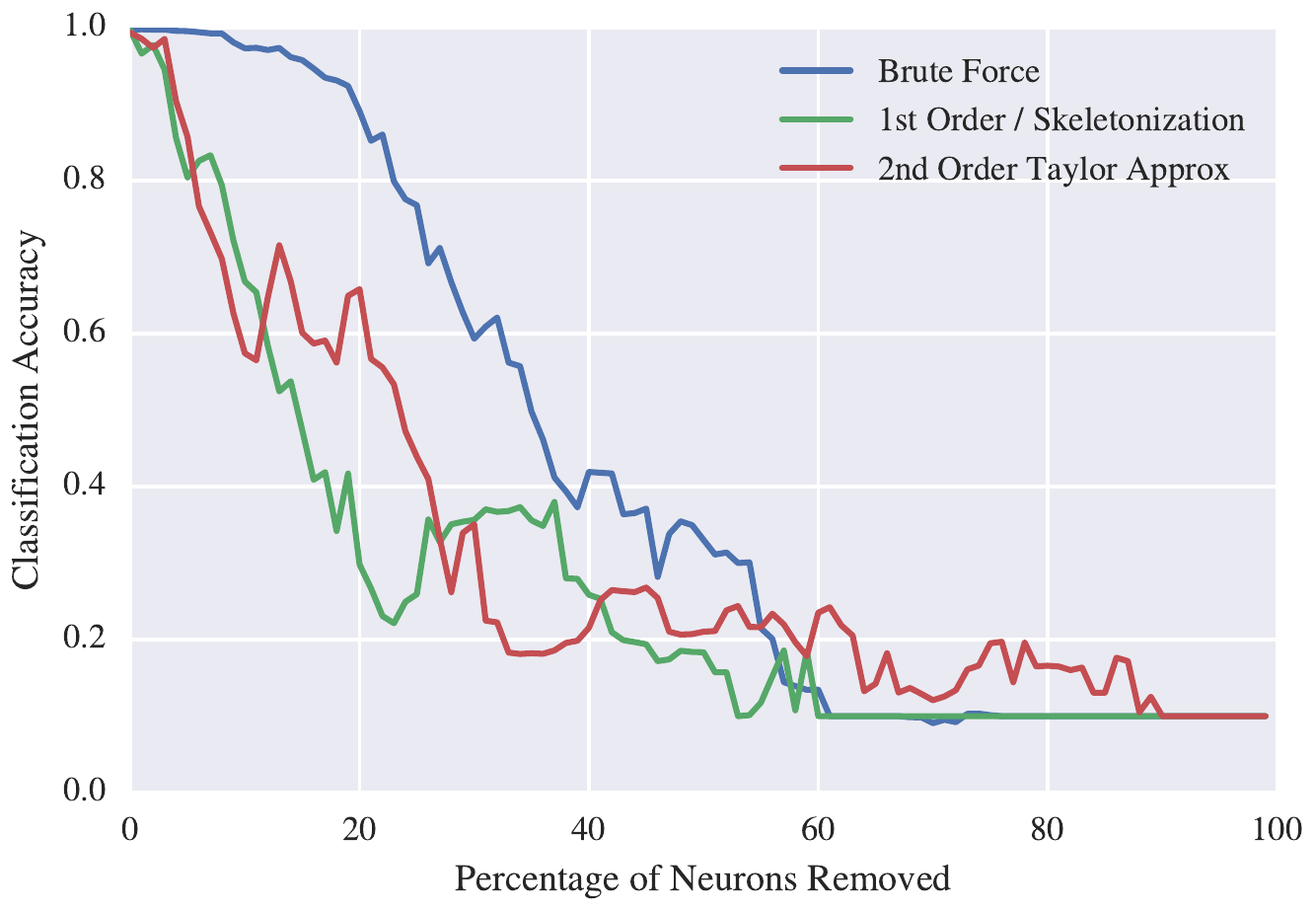}
\caption{Degradation in squared error (left) and classification accuracy (right) after pruning a 2-layer network using the Single Overall Ranking algorithm; (\textbf{Network:} 2 layers, 50 neurons/layer, 10 outputs, logistic sigmoid activation, starting test accuracy: 1.000)}
\label{fig:mnist-single-ranking-double-layer}
\end{figure}

Figure \ref{fig:mnist-single-ranking-double-layer} shows the pruning results for Algorithm \ref{algo1} on a 2-layer network. The ranking procedure is identical to the one used to generate Figure \ref{fig:mnist-single-ranking-single-layer}. 

Unsurprisingly, a 2-layer network is harder to prune because a single overall ranking will never capture the interdependencies between neurons in different layers. It makes sense that this is worse than the performance on the 1-layer network, even if this method is already known to be bad, and we'd likely never use it in practice. 

\subsubsection{iterative re-ranking Algorithm}
\begin{figure}[!ht]
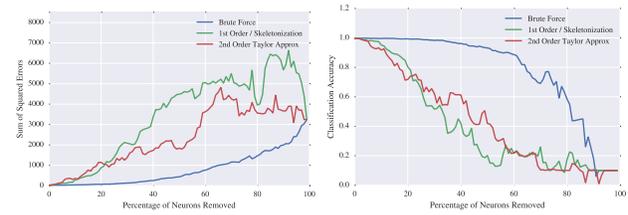

\centering
\includegraphics[width=0.49\linewidth]{mnist-deep-iterative-rerank-method.pdf}
\includegraphics[width=0.49\linewidth]{mnist-deep-iterative-rerank-accuracy.pdf}
\caption{Degradation in squared error (left) and classification accuracy (right) after pruning a 2-layer network using the iterative re-ranking algorithm; (\textbf{Network:} 2 layers, 50 neurons/layer, 10 outputs, logistic sigmoid activation, starting test accuracy: 1.000)}
\label{fig:mnist-re-ranking-double-layer}
\end{figure}

Figure \ref{fig:mnist-re-ranking-double-layer} shows the results from using Algorithm \ref{algo2} on a 2-layer network. We compute the same brute force rankings and Taylor series approximations of error deltas over the remaining active neurons in the network after each pruning decision used to generate Figure \ref{fig:mnist-re-ranking-single-layer}. Again, this is intended to account for the effects of canceling interactions between neurons.

\subsubsection{Visualization of Error Surface \& Pruning Decisions}
As seen in the case of a single layered network, these graphs are a visualization the error surface of the network output with respect to the neurons chosen for removal using each algorithm, represented in intervals of 10 neurons. 

\subsubsection{Visualization of brute force Pruning Decisions}
\begin{figure}[!ht]
\centering
\includegraphics[width=\linewidth]{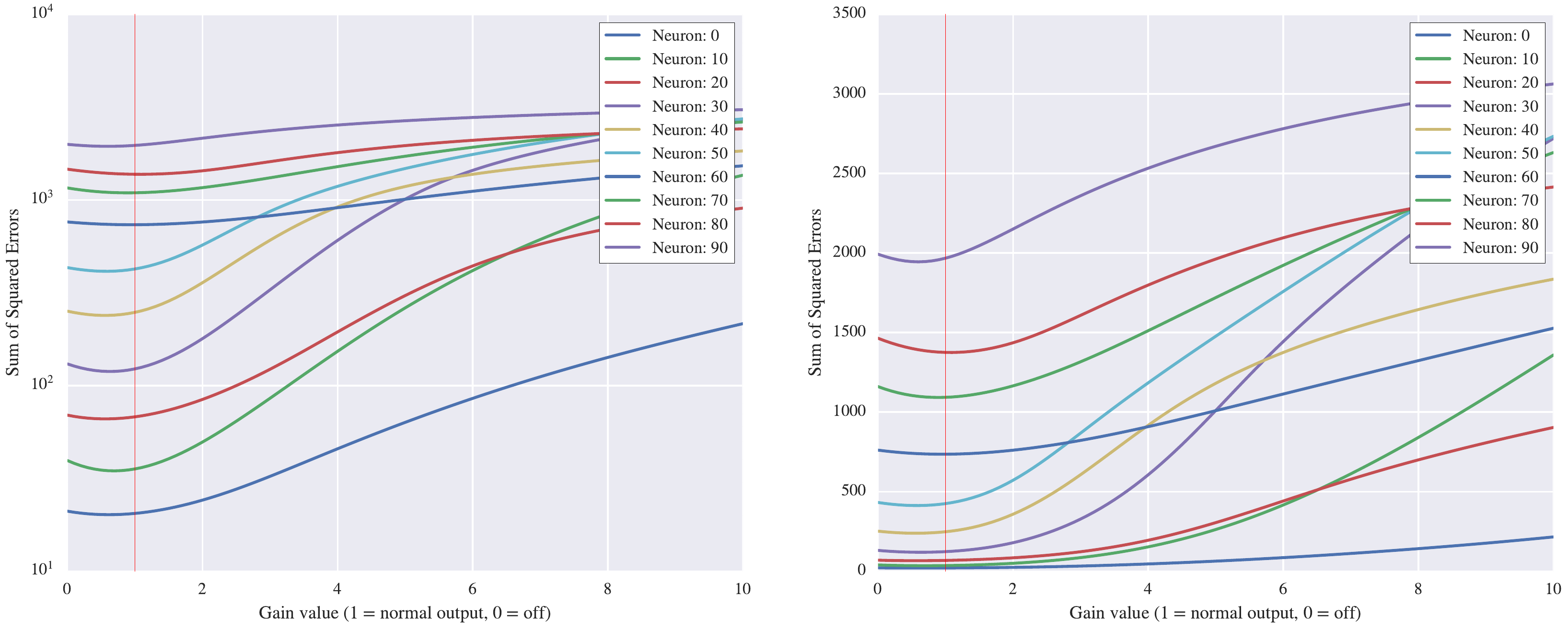}
\caption{Error surface of the network output in log space (left) and real space (right) with respect to each candidate neuron chosen for removal using the brute force criterion; (\textbf{Network:} 2 layers, 50 neurons/layer, 10 outputs, logistic sigmoid activation, starting test accuracy: 1.000)}
\label{fig:mnist-gt-double-layer}
\end{figure}

In Figure \ref{fig:mnist-gt-double-layer}, it is clear why these neurons got chosen, their graphs clearly show little change when neuron is removed, are mostly near the floor, and show convex behaviour of error surface, which argues for the rationalization of using 2nd order methods to estimate difference in error when they are turned off.

\subsubsection{Visualization of 1st Order Approximation Pruning Decisions}
\begin{figure}[!ht]
\centering
\includegraphics[width=\linewidth]{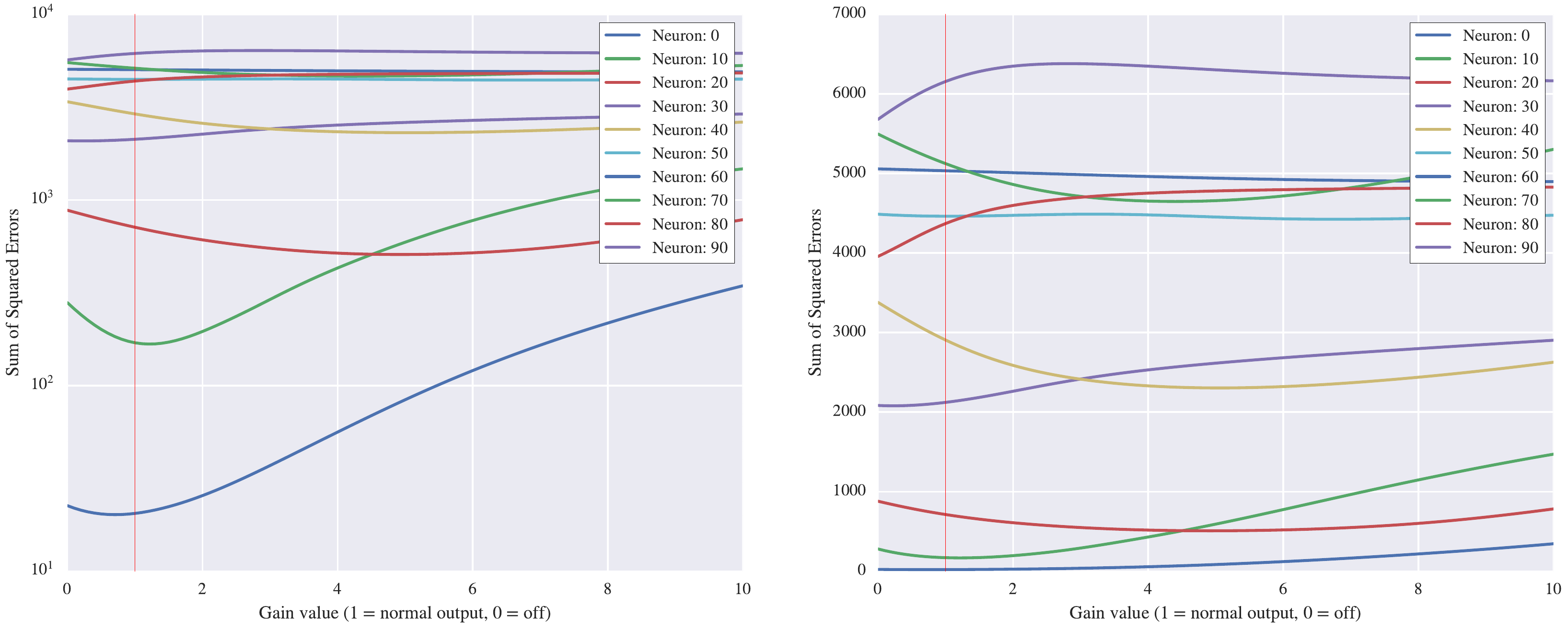}
\caption{Error surface of the network output in log space (left) and real space (right) with respect to each candidate neuron chosen for removal using the 1st order Taylor Series error approximation criterion; (\textbf{Network:} 2 layers, 50 neurons/layer, 10 outputs, logistic sigmoid activation, starting test accuracy: 1.000)}
\label{fig:mnist-g1-double-layer}
\end{figure}

Drawing a flat line at the point of each neurons intersection with the red vertical line (no change in gain) shows that the 1st derivative method is actually accurate for estimation of change in error in these cases, but still ultimately leads to poor decisions. 

\subsubsection{Visualization of 2nd Order Approximation Pruning Decisions}
\begin{figure}[!ht]
\centering
\includegraphics[width=\linewidth]{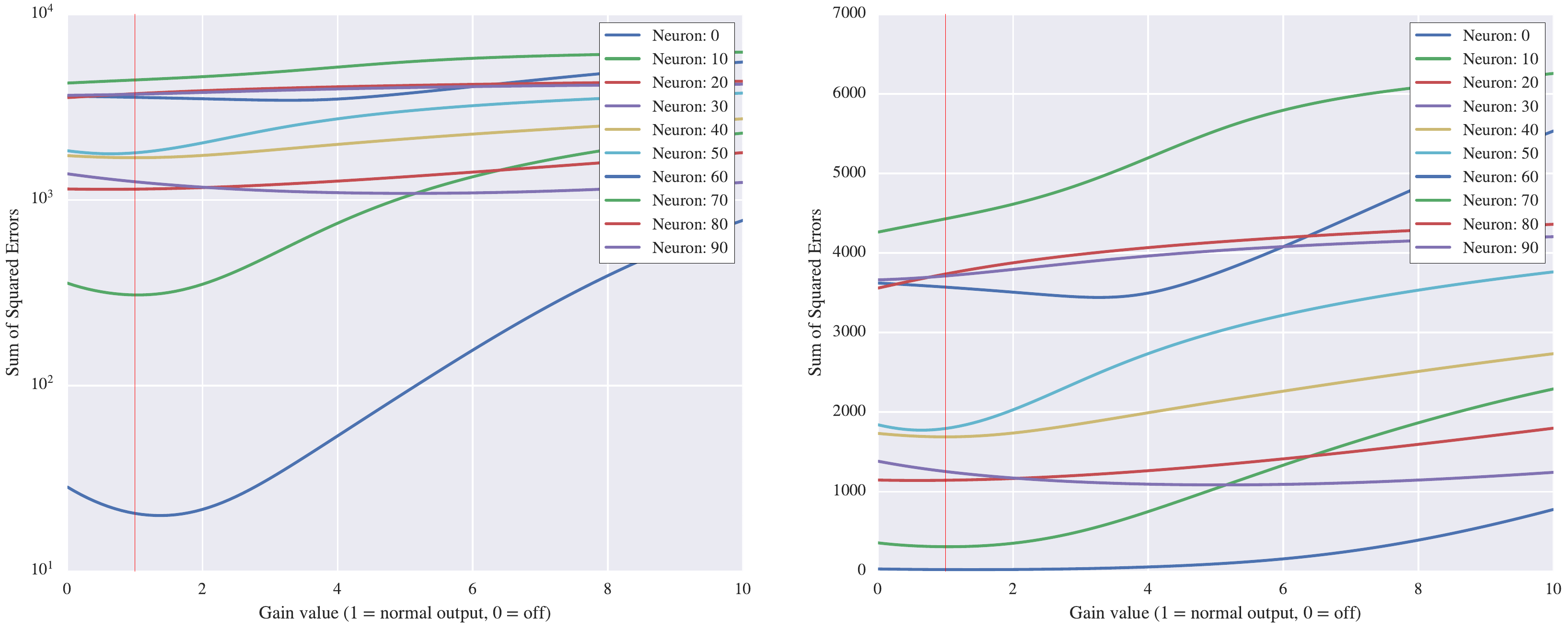}
\caption{Error surface of the network output in log space (left) and real space (right) with respect to each candidate neuron chosen for removal using the 2nd order Taylor Series error approximation criterion; (\textbf{Network:} 2 layers, 50 neurons/layer, 10 outputs, logistic sigmoid activation, starting test accuracy: 1.000)}
\label{fig:mnist-g2-double-layer}
\end{figure}

Clearly these neurons are not overtly poor candidates for removal (error doesn't change much between 1.0 \& zero-crossing left-hand-side), but could be better (as described above in the brute force Criterion discussion).

\subsection{Experiments on Toy Datasets}

As can be seen from the experiments on MNIST, even though the 2nd-order approximation criterion is consistently better than 1st-order, its performance is not nearly as good as brute force based ranking, especially beyond the first layer. What is interesting to note is that from some other experiments conducted on toy datasets (predicting whether a given point would lie inside a given shape on the Cartesian plane), the performance of the 2nd-order method was found to be exceptionally good and produced results very close to the brute force method. The 1st-order method, as expected, performed poorly here as well. Some of these results are illustrated in Figure \ref{fig:diamond}. 

\begin{figure}[!ht]
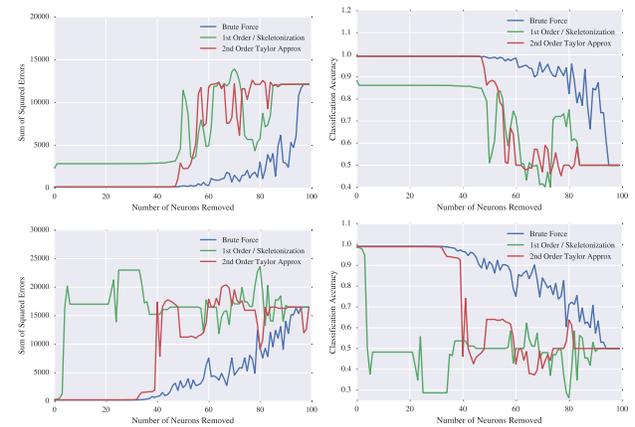

\centering
\includegraphics[width=0.49\linewidth]{diamond-iterative-rerank-method.pdf}
\includegraphics[width=0.49\linewidth]{diamond-iterative-rerank-accuracy.pdf}
\includegraphics[width=0.49\linewidth]{rshape-iterative-rerank-method.pdf}
\includegraphics[width=0.49\linewidth]{rshape-iterative-rerank-accuracy.pdf}
\caption{Degradation in squared error (left) and classification accuracy (right) after pruning a 2-layer network using the iterative re-ranking algorithm on a toy ``diamond'' shape dataset (top) and a toy ``random shape'' dataset (below); (\textbf{Network:} 2 layers, 50 neurons/layer, 10 outputs, logistic sigmoid activation, starting test accuracy: 0.992(diamond); 0.986(random shape)}
\label{fig:diamond}
\end{figure}


\end{document}